\documentclass[11pt]{article}

\usepackage[final]{acl}
\usepackage{xcolor}
\usepackage{colortbl}
\definecolor{lightblue}{RGB}{235,245,255}
\usepackage{times}
\usepackage{booktabs}
\usepackage[table]{xcolor}
\definecolor{groupgray}{gray}{0.93}
\usepackage{amsthm}
\newtheorem{proposition}{Proposition}
\usepackage{latexsym}
\usepackage{amsmath} 
\usepackage{amssymb,amsfonts}
\usepackage[T1]{fontenc}
\usepackage{booktabs}

\usepackage[utf8]{inputenc}
\usepackage{svg}

\usepackage{microtype}

\usepackage{inconsolata}

\usepackage{amsthm}
\usepackage{adjustbox} 
\usepackage{graphicx}
\usepackage{caption}
\usepackage{booktabs}
\usepackage{makecell}
\usepackage{array}
\usepackage{placeins} 
\usepackage{multirow}
\usepackage{titletoc}
\newtheorem{theorem}{Theorem}[section] 
\usepackage[most]{tcolorbox}
\usepackage{xcolor}
\usepackage{subcaption}
\usepackage{caption}
\usepackage[most]{tcolorbox}
\tcbuselibrary{skins,breakable,listings}

\definecolor{BraceBlue}{RGB}{80,180,255} 
\usepackage{etoc}
\tcbset{
  promptboxBase/.style={
    enhanced,
    sharp corners,
    boxrule=0.8pt,
    left=1mm,right=1mm,top=0.5mm,bottom=0.5mm,
    fonttitle=\bfseries,
    attach boxed title to top left={xshift=1.5mm,yshift=-1mm},
    boxed title style={boxrule=0pt, sharp corners, width=\dimexpr\linewidth-6mm\relax},
  }
}

\usepackage{listings}

\tcbset{
  promptboxBase/.style={
    enhanced,
    sharp corners,
    boxrule=0.8pt,
    left=1mm,right=1mm,top=0.5mm,bottom=0.5mm,
    fonttitle=\bfseries,
    attach boxed title to top left={xshift=1.5mm,yshift=-1mm},
  }
}

\tcbset{
  promptboxA/.style={promptboxBase, colback=blue!4,  colframe=blue!55!black,  coltitle=blue!55!black,
    boxed title style={boxrule=0pt, sharp corners, width=\dimexpr\linewidth-8mm\relax, colback=blue!10,  colframe=blue!10}},
  promptboxB/.style={promptboxBase, colback=green!4, colframe=green!50!black, coltitle=green!50!black,
    boxed title style={boxrule=0pt, sharp corners, width=\dimexpr\linewidth-8mm\relax, colback=green!10, colframe=green!10}},
  promptboxC/.style={promptboxBase, colback=orange!5,colframe=orange!70!black,coltitle=orange!70!black,
    boxed title style={boxrule=0pt, sharp corners, width=\dimexpr\linewidth-8mm\relax, colback=orange!12,colframe=orange!12}},
  promptboxD/.style={promptboxBase, colback=purple!5,colframe=purple!55!black,coltitle=purple!55!black,
    boxed title style={boxrule=0pt, sharp corners, width=\dimexpr\linewidth-8mm\relax, colback=purple!12,colframe=purple!12}},
  promptboxE/.style={promptboxBase, colback=teal!4,  colframe=teal!55!black,  coltitle=teal!55!black,
    boxed title style={boxrule=0pt, sharp corners, width=\dimexpr\linewidth-8mm\relax, colback=teal!10,  colframe=teal!10}},
}

\newtcblisting{PromptBoxA}[2][]{promptboxA, listing only,
  title={\parbox[t]{\dimexpr\linewidth-8mm\relax}{\raggedright\sloppy\textbf{#2}}},
  listing options={basicstyle=\small,breaklines=true, columns=fullflexible, keepspaces=true, showstringspaces=false}, #1}

\newtcblisting{PromptBoxB}[2][]{promptboxB, listing only,
  title={\parbox[t]{\dimexpr\linewidth-8mm\relax}{\raggedright\sloppy\textbf{#2}}},
  listing options={basicstyle=\small,breaklines=true, columns=fullflexible, keepspaces=true, showstringspaces=false}, #1}

\newtcblisting{PromptBoxC}[2][]{promptboxC, listing only,
  title={\parbox[t]{\dimexpr\linewidth-8mm\relax}{\raggedright\sloppy\textbf{#2}}},
  listing options={basicstyle=\small,breaklines=true, columns=fullflexible, keepspaces=true, showstringspaces=false}, #1}

\newtcblisting{PromptBoxD}[2][]{promptboxD, listing only,
  title={\parbox[t]{\dimexpr\linewidth-8mm\relax}{\raggedright\sloppy\textbf{#2}}},
  listing options={basicstyle=\small,breaklines=true, columns=fullflexible, keepspaces=true, showstringspaces=false}, #1}

\newtcblisting{PromptBoxE}[2][]{promptboxE, listing only,
  title={\parbox[t]{\dimexpr\linewidth-8mm\relax}{\raggedright\sloppy\textbf{#2}}},
  listing options={basicstyle=\small,breaklines=true, columns=fullflexible, keepspaces=true, showstringspaces=false}, #1}
\usepackage[ruled,vlined,linesnumbered]{algorithm2e}
\title{CAMO: An Agentic Framework for Automated Causal Discovery from Micro Behaviors to Macro Emergence in LLM Agent Simulations}

\author{
\textbf{Xiangning Yu\textsuperscript{1,2,3}\thanks{Equal contribution}},
\textbf{Yuwei Guo\textsuperscript{1,2,3}\footnotemark[1]},
\textbf{Yuqi Hou\textsuperscript{1,2,3}},
\textbf{Xiao Xue\textsuperscript{1,2,3}\thanks{Corresponding author}},
\textbf{Qun Ma\textsuperscript{1,2,3}}
\\[0.5em]
\textsuperscript{1}College of Intelligence and Computing, Tianjin University, Tianjin, China\\
\textsuperscript{2}Tianjin Key Laboratory of Healthy Habitat and Smart Technology, Tianjin, China\\
\textsuperscript{3}Laboratory of Computation and Analytics of Complex Management Systems,\\ Tianjin University, Tianjin, China
\\[0.5em]
\texttt{\{yxn9191, 2024244171, houyuqi, jzxuexiao, 1023244018\}@tju.edu.cn}
}

\begin{document}
\maketitle

\begin{abstract}
LLM-empowered agent simulations are increasingly used to study social emergence, yet the micro-to-macro causal mechanisms behind macro outcomes often remain unclear. This is challenging because emergence arises from intertwined agent interactions and meso-level feedback and nonlinearity, making generative mechanisms hard to disentangle. To this end, we introduce \textbf{\textsc{CAMO}}, an automated \textbf{Ca}usal discovery framework from \textbf{M}icr\textbf{o} behaviors to \textbf{M}acr\textbf{o} Emergence in LLM agent simulations. \textsc{CAMO} converts mechanistic hypotheses into computable factors grounded in simulation records and learns a compact causal representation centered on an emergent target $Y$. \textsc{CAMO} outputs a computable Markov boundary and a minimal upstream explanatory subgraph, yielding interpretable causal chains and actionable intervention levers. It also uses simulator-internal counterfactual probing to orient ambiguous edges and revise hypotheses when evidence contradicts the current view. Experiments across four emergent settings demonstrate the promise of \textsc{CAMO}.\footnote{The code is available at: \url{https://github.com/RisingDate/CAMO}.}
\end{abstract}

\section{Introduction}
\label{sec:intro}

LLM-empowered agent simulations have emerged as a powerful laboratory for
studying complex social phenomena~\cite{park2023generative,piao2025agentsociety,xue2023chatgpt,ma2024computational}.
By instantiating populations of agents that make autonomous, LLM-driven
decisions, these simulations generate rich, adaptive, and often non-linear
interaction patterns. Through decentralized interactions, macro-level emergent
outcomes such as coordination, norm formation, and polarization naturally arise
\cite{riedl2025emergent,chen2024agentverse,ren2024emergence,takata2024spontaneous,
piatti2024cooperate}.

Although emergent patterns are frequently observed in such simulations, they
provide limited insight into the causal mechanisms that generate them
\cite{mou2024individual,guo2024large,piao2025emergence,yu2026invariant}. In particular, it often remains unclear
how micro-level agent behaviors and meso-level interaction structures jointly
give rise to a target emergent outcome, making interventions on prompts, agent
policies, or interaction settings largely heuristic and difficult to
generalize~\cite{xue2023computational2,xue2024computational1,xue2024computational3,xue2021soa}.

\begin{figure}[t]
    \centering
    \includegraphics[width=\columnwidth]{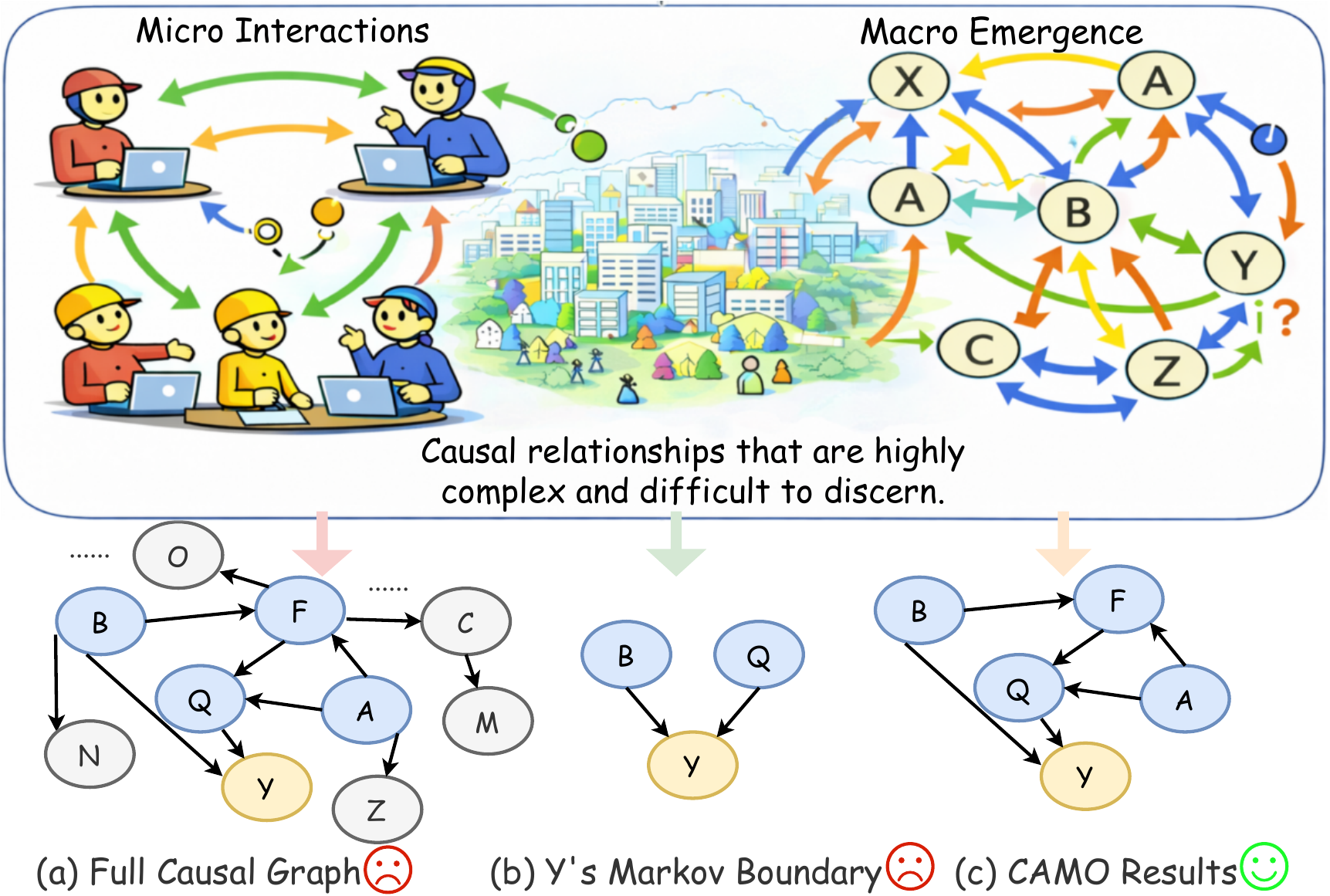}
    \caption{\textbf{Causal representations recovered by \textsc{CAMO}.}
    \textsc{CAMO} identifies a compact causal neighborhood around the target outcome $Y$
    that is sufficient for causal identification, and augments it with a minimal
    set of upstream pathways needed to explain and support intervention on
    micro-to-macro emergence.}
    \label{fig:CAMO_overview}
\end{figure}

Recent work has explored the integration of large language models with causal
discovery and causal reasoning
\citep{jiralerspong2024efficient,le2024multi,takata2024spontaneous,
kiciman2023causal,zevcevic2023causal}, but these methods are largely developed for
static variable spaces and do not address how causal mechanisms produce emergent
outcomes in adaptive multi-agent systems. In parallel, LLM-empowered agent
simulations have been widely used to study emergence phenomena
\citep{mao2025agent, park2023generative,ren2024emergence,gupta2025role,chuang2024simulating,xiao2023putational4,xue2023computational5},
yet they primarily demonstrate or characterize emergent behavior rather than
uncovering the causal mechanisms by which it arises. Consequently, the
micro-to-macro causal structures needed to explain and reliably intervene on a
specific emergent outcome remain largely uncharacterized.

In this work, we ask: \emph{given an LLM-empowered agent simulation, how can we
automatically recover the causal mechanisms that link micro-level agent
behaviors and meso-level interactions to an emergent macro-level outcome $Y$, in
a way that supports both explanation and intervention?}

We introduce \textsc{CAMO}, a multi-agent framework for micro-to-macro causal
discovery in LLM-empowered agent simulations. Rather than attempting to recover
the full causal structure underlying emergence, which is often infeasible and
unnecessary in complex agent-based systems, \textsc{CAMO} identifies a
\emph{minimal yet sufficient} set of upstream variables and causal pathways that
explain a given emergent outcome $Y$.

The central idea of \textsc{CAMO} is to explicitly separate \emph{causal identification}
from \emph{mechanistic explanation}. Causal identification focuses on recovering
a minimal local causal structure around $Y$ that supports prediction and
intervention. Mechanistic explanation then traces this structure backward
through upstream micro- and meso-level pathways, yielding an interpretable
account of how emergence arises. This separation allows \textsc{CAMO} to remain
statistically grounded while preserving clear micro-to-macro causal narratives.

Operationally, \textsc{CAMO} treats the LLMs driving agent behavior inside the
simulation as a black-box stochastic data-generating process. A separate set of
\emph{workflow LLMs} operates outside the simulation to interpret textual
descriptions, propose and revise mechanistic hypotheses, and coordinate
discovery. Candidate causal relations are admitted, pruned, and oriented using
constraints from observational records and targeted simulator-internal
counterfactual interventions, organized through a fast--slow self-evolution
loop. 

Our contributions are threefold:
\begin{itemize}
    \item We frame micro-to-macro causal discovery in LLM-empowered agent
    simulations as the problem of identifying a minimal but sufficient upstream
    causal explanation for an emergent outcome.
\item We propose \textsc{CAMO}, the first multi-agent framework for studying \emph{micro-to-macro emergence mechanisms} in LLM-based simulations that combines domain priors, observational data, and simulator-internal counterfactuals.
\item We introduce a fast--slow self-evolving loop that curbs hallucinated priors and noisy interventions, improving robustness without assuming global identifiability.
\end{itemize}
\section{Related Work}
\label{sec:related-work}

\paragraph{LLM-assisted causal discovery.}
Recent work explores using large language models to assist causal discovery and
causal reasoning, including proposing causal relations, constraints, or latent
factors to complement statistical methods
\citep{chi2024unveiling,jin2023cladder,liu2025large,kiciman2023causal,wu2024causality,feng2024reliability}.
Several approaches integrate LLM-generated knowledge into classical structure
learning pipelines under limited or noisy data
\citep{takayama2024integrating,jiralerspong2024efficient,farooq2023understanding}.
More recent systems formulate causal discovery as an agentic or tool-augmented
workflow, including multi-agent refinement and autonomous causal modeling
\citep{le2024multi,liu2024discovery,
han2024causal,khatibi2025alcmautonomousllmaugmentedcausal}.
These methods are primarily designed for fixed or predefined variable spaces and
do not address causal discovery in adaptive, emergent multi-agent environments
\citep{richens2024robust,yu2025causal,yu2018mining}.

\paragraph{LLM-based simulation and emergence.}
LLM-empowered agent simulations have been widely used to study emergent social
phenomena and collective dynamics
\citep{mao2025agent,park2023generative,gao2024large,ashery2025emergent,anthis2025llm}.
Large-scale platforms demonstrate that thousands of interacting LLM agents can
exhibit coordination, cooperation, and polarization
\citep{gurcan2024llm,piao2025agentsociety, piao2025emergence, ren2024emergence,piatti2024cooperate}.
At the individual level, generative agent simulations validate the behavioral
fidelity of LLM agents
\citep{agrawal2024vidur,adornetto2025generative,chuang2024simulating,dai2024artificial}.
However, existing studies primarily describe or reproduce emergent patterns,
while the underlying micro-to-macro causal mechanisms remain largely unexplored
from a causal view
\citep{zhang2025socioverseworldmodelsocial,wang2025limits,xia2024llm}.
\begin{figure*}[t]
\centering
\includegraphics[width=1\textwidth]{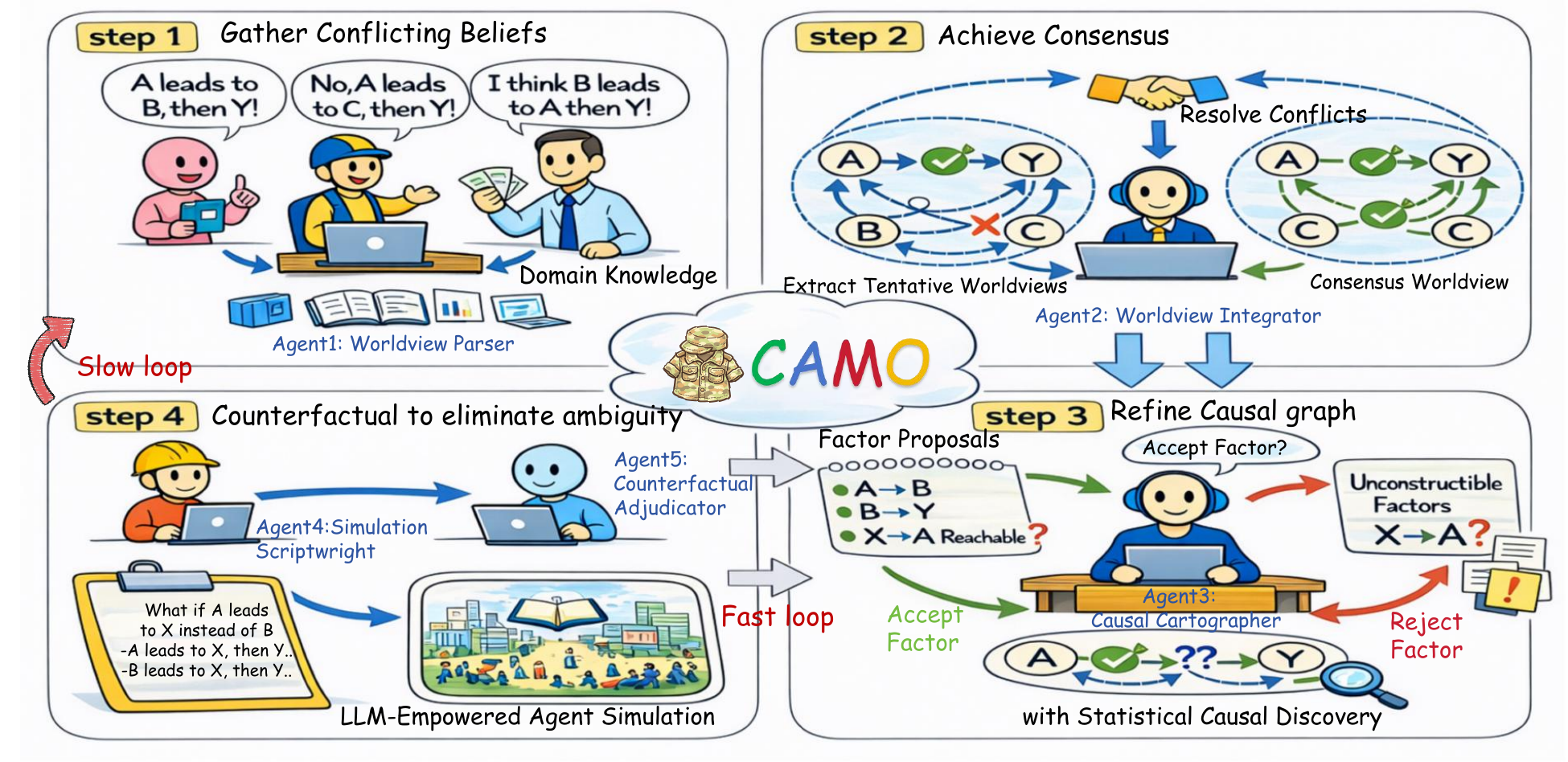}
\caption{\textbf{Overview of \textsc{CAMO}.}
A fast--slow loop integrates textual worldviews, causal discovery, and
simulator-internal interventions to recover a minimal causal interface and
micro-to-macro explanation for the target outcome.}
\label{fig:zhutu}
\vspace{-1mm}
\end{figure*}

\begin{algorithm}[t]
\caption{\textsc{CAMO}: Local Interface + Minimal Emergence Explanation}
\label{alg:CAMO}
\KwIn{Simulator $\mathcal{S}$, target $Y$, observations $X_{\mathrm{obs}}$, user query $Q$}
\KwOut{$\mathrm{MB}^{\mathcal H}(Y)$ and $E_Y$}

\textbf{A1:} Retrieve domain texts $T$ from $Q$; parse $T$ into worldviews $\{W_a\}$\;

\Repeat(\tcp*[f]{Slow loop}){$\mathrm{MB}^{\mathcal H}(Y)$ and $E_Y$ stabilize}{
  \textbf{A2:} Select decision worldview $W$ and mechanism graph $G_W$\;

  \Repeat(\tcp*[f]{Fast loop}){local graph stabilizes or counterfactuals become uninformative}{
    \textbf{A3:} Add/prune factors $\rightarrow$ stabilize $V$ and $\mathrm{MB}^{\mathcal H}(Y)$\;
    \textbf{A3:} Constrained discovery $\rightarrow$ CPDAG/PAG; rank ambiguities\;
    \textbf{A4--A5:} Counterfactual queries; add confirmed constraints\;
    \textbf{A3:} Update local graph\;
  }

  \If{counterfactual evidence contradicts $W$}{Update/remove $W$; \textbf{continue}\;}
  \textbf{A3:} Extract $E_Y$ from $G_W$ s.t.\ $\mathcal{R}\Rightarrow \mathrm{MB}^{\mathcal H}(Y)$\;
}
\end{algorithm}

\section{Methodology}
\label{sec:method}

\subsection{Problem Formulation and Objectives}
\label{sec:method:problem}

We study causal discovery for emergent outcomes in \emph{LLM-empowered agent
simulations}, where populations of agents interact and collectively give rise to a macro-level target
variable $Y$.
Let $X_{\mathrm{obs}}$ denote the full set of logged simulation observables.
Beyond logged signals, \textsc{CAMO} introduces additional variables as
explicit, \emph{computable} \emph{nonlinear} transforms of $X_{\mathrm{obs}}$,
since many micro/meso mechanisms are not primitive logs but only become
measurable through such compositions, yielding an induced factor space $\mathcal{H}$ built from a fixed set of
interpretable templates (e.g., ratios, rolling statistics, and graph metrics).

Rather than recovering the full simulator causal DAG, we focus on a validated
\emph{local causal interface} around $Y$ and a minimal mechanistic explanation
that traces emergence from micro/meso-level causes to $Y$.
Formally, \textsc{CAMO} outputs:
\begin{equation}
\label{eq:CAMO_outputs}
\Big(\ \mathrm{MB}^{\mathcal H}(Y),\ E_Y\ \Big).
\end{equation}
Here, $\mathrm{MB}^{\mathcal H}(Y)$ denotes the Markov boundary of $Y$ in the
induced factor space, i.e., the minimal sufficient conditioning set for
predicting and intervening on $Y$, while the explanatory subgraph $E_Y$ augments
this local causal interface with a minimal set of upstream micro- and
meso-level variables and pathways, providing a mechanistic account of how
emergence arises.

\subsection{Overview of \textsc{CAMO}}
\label{sec:method:overview}

\textsc{CAMO} is a multi-agent framework consisting of five LLM agents:
the Worldview Parser (A1), Worldview Integrator (A2), Causal Cartographer (A3),
Simulation Scriptwright (A4), and Counterfactual Adjudicator (A5).
These agents coordinate in two loops: a fast refinement loop that proposes and prunes computable factors to fit a local causal model under a fixed hypothesis,
followed by targeted interventions to resolve remaining ambiguities.
A slow revision loop is triggered when interventional evidence contradicts the hypothesis, prompting hypothesis updates.
Algorithm~\ref{alg:CAMO} summarizes the procedure; Figure~\ref{fig:zhutu} illustrates it.

\paragraph{A1: Worldview Parser.}
As a \textsc{CAMO} workflow agent, A1 converts unstructured domain knowledge into structured
mechanistic causal hypotheses. Given a user query $Q$ specifying the simulator setting and target
outcome $Y$, A1 automatically retrieves a relevant text bundle $T$ (e.g., domain references and
background materials) and parses $T$ into candidate variables and relations. It merges semantically
equivalent mentions, tags variables by micro/meso/macro scale, and preserves conflicts from
heterogeneous sources (e.g., opposite directions under different assumptions) as explicit competing
alternatives. The result is a set of revisable causal worldviews that serve as priors for downstream
discovery.

\paragraph{A2: Worldview Integrator.}
A2 aligns the perspective-indexed worldviews into a shared, \emph{computable}
representation.
It (i) unifies semantically equivalent variables, (ii) assigns canonical
construction rules for computable factors (e.g., ratios, summary statistics,
graph metrics), and (iii) makes cross-perspective conflicts explicit.
A2 selects a decision worldview $W^{(k)}$ for the current refinement round (see Appendix~\ref{app:judge} for selection criteria), and
also produces a retained \emph{mechanism graph} $G_W^{(k)}$ that stores upstream
causal pathways and alternative hypotheses for later revision.

\subsection{Local Interface Learning by Refinement (A3: Causal Cartographer)}
\label{sec:method:a3}

\paragraph{Data-grounded representation.}
Let $V^{(t)}$ be the representation variables retained at refinement step $t$,
where $V^{(t)} \subseteq \mathcal H$ consists of variables that are either
directly logged in $X_{\mathrm{obs}}$ or computably constructed from
$X_{\mathrm{obs}}$.

\paragraph{Add--prune refinement.}
Under the current decision worldview, A3 refines $V^{(t)}$ by testing each
computable factor $Z$ via its estimated information gain for $Y$:
\begin{equation}
\label{eq:add_rule}
\widehat{\Delta I}(Z)
:= \widehat H(Y\mid V^{(t)})-\widehat H(Y\mid V^{(t)},Z) \;>\; \tau_I .
\end{equation}
That is, $Z$ is added only if it reduces the conditional uncertainty of $Y$ by
at least $\tau_I$. A3 then prunes any factor that becomes conditionally
redundant for predicting $Y$ given the remaining variables in $\mathcal H$.
This add--prune loop yields a compact \emph{computable Markov boundary}
$\mathrm{MB}^{\mathcal H}(Y)$; see Appendix~\ref{app:pc_estimation}
for estimation details.

\paragraph{Constrained local discovery.}
Given the stabilized representation $V^{(t)}$, A3 performs constraint-based causal
discovery under the structural constraints specified by $W^{(k)}$, selecting a
suitable test/algorithm based on the data regime, and outputs a partially
oriented equivalence-class graph (e.g., a CPDAG or PAG). Edges whose orientation
remains ambiguous are prioritized (by estimated effect importance and orientation
uncertainty) and forwarded to A4 for intervention design.

\subsection{Minimal Connecting Explanatory Subgraph (A3: Causal Cartographer)}
\label{sec:method:ees}

\paragraph{Root variables.}
We define root variables $\mathcal{R}$ as upstream variables in the retained
mechanism graph $G_W^{(k)}$ that correspond to fundamental agent behaviors or
environment settings, and serve as anchors for tracing micro-to-macro emergence.

\paragraph{Definition of $E_Y$ via minimal connecting subgraph.}
After stabilizing the computable Markov boundary $\mathrm{MB}^{\mathcal H}(Y)$
and incorporating simulation-confirmed constraints, A3 constructs an
explanatory subgraph
\begin{equation}
\label{eq:ees_def}
\begin{aligned}
E_Y \;:=\;& \{Y\}\cup \mathrm{MB}^{\mathcal H}(Y) \\
&\cup\;\mathrm{Conn}_{\min}\!\Big(\mathcal{R}\Rightarrow \mathrm{MB}^{\mathcal H}(Y)\Big).
\end{aligned}
\end{equation}
where $\mathrm{Conn}_{\min}(\mathcal{R}\Rightarrow \mathrm{MB}^{\mathcal H}(Y))$
denotes a connecting subgraph of $G_W^{(k)}$ that, for each boundary variable
$B\in \mathrm{MB}^{\mathcal H}(Y)$, retains all nodes and edges that lie on
directed paths from some root $r\in\mathcal{R}$ to $B$, while enforcing all
simulation-confirmed edges as hard constraints.
It is \emph{minimal}: removing any retained node/edge would break the required
$r\!\to\!B$ reachability for some $B\in \mathrm{MB}^{\mathcal H}(Y)$. Computation
details are in Appendix~\ref{sec:appendix:conn_min}.

Intuitively, $\mathrm{MB}^{\mathcal H}(Y)$ provides a minimal \emph{local causal
interface} sufficient for identifying and intervening on $Y$, while $E_Y$
augments this interface with the minimal set of upstream micro- and meso-level
causal pathways needed to explain how the emergent outcome arises.

\subsection{Interventions and Counterfactual Evidence (A4--A5)}
\label{sec:method:a4a5}

\paragraph{A4: Simulation Scriptwright.}
A4 converts prioritized ambiguous edges into executable intervention scripts
when the simulator admits a realizable control (e.g., prompt/policy override or
environment parameter change); prioritization uses a computable importance--uncertainty
score (Appendix~\ref{app:a4_ucb}).
Each script specifies the intervention endpoint, levels, paired-rollout
protocol holding other randomness fixed, and replication budget across
configurations.

\paragraph{A5: Counterfactual Adjudicator.}
A5 executes paired interventions (e.g., $\mathrm{do}(X{=}x)$ vs.\ $\mathrm{do}(X{=}x')$)
to obtain simulator-internal counterfactual contrasts. An edge is marked
\emph{simulation-confirmed} if effects are consistently nonzero and orientations
are consistent across configurations. Confirmed edges are treated as strong
constraints in subsequent fast-loop updates.

\section{Theoretical Analysis}
\label{sec:theory}

This section analyzes theoretical properties of \textsc{CAMO}
(\textsection~\ref{sec:method}). Proofs are given in Appendix~\ref{app:proofs}.

\subsection{Convergence to a Computable Markov Boundary}
\label{sec:theory:mb}

Let $V^{(t)}$ denote the representation variables available at refinement step
$t$, including both observed variables and computably constructed factors in
the induced factor space $\mathcal H$. Let $B_t$ denote the Markov boundary of
$Y$ within $V^{(t)}$ \citep{koller2009probabilistic}.

\begin{theorem}[One-step compactness of boundary refinement]
\label{thm:monotone}
Under the add--prune refinement procedure of Agent~A3, suppose refinement step
$t\!\to\!t{+}1$ admits a (possibly empty) set of new factors $\mathcal Z_{t+1}$
with $m_t := |\mathcal Z_{t+1}|$. Then the Markov boundary size satisfies
\begin{equation}
\label{eq:monotone}
|B_{t+1}| \le |B_t| + m_t,
\end{equation}
with strict decrease possible whenever some previously boundary variable is
pruned as conditionally redundant given the remaining variables.
\end{theorem}

Following~\citet{liu2024discovery}, we characterize convergence beyond boundary
size by measuring the remaining unexplained dependence between $Y$ and the full
observation space:
\begin{equation}
\label{eq:Ft}
F_t := I(Y;X_{\mathrm{obs}} \mid V^{(t)}).
\end{equation}

\begin{theorem}[Geometric decay under refinement capability]
\label{thm:geom}
Assume that with probability at least $p>0$, a refinement step admits at least one
informative computable factor that reduces the residual dependence, and that
conditioned on such a step the reduction is by a constant fraction $C>0$, i.e.,
\begin{equation}
\label{eq:contraction}
F_{t+1} \le (1-C)F_t ,
\end{equation}
and otherwise $F_{t+1}\le F_t$. Then the expected residual dependence decays
geometrically:
\begin{equation}
\label{eq:geom}
\mathbb{E}[F_t] \le (1-pC)^t F_0.
\end{equation}
Consequently, refinement stabilizes (up to equivalence in the induced factor space $\mathcal H$) on a
$Y$-sufficient representation, which can be pruned to yield a computable Markov boundary
$\mathrm{MB}^{\mathcal H}(Y)$.
\end{theorem}

The constants $p$ and $C$ are used only to characterize refinement effectiveness;
they are estimated a posteriori from the $\{F_t\}$ trajectory
(Appendix~\ref{app:pc_estimation}) and are not required by the algorithm.

Theorem~\ref{thm:monotone} ensures the computable Markov boundary cannot expand
uncontrollably, while Theorem~\ref{thm:geom} implies refinement converges (up to
equivalence in $\mathcal H$) to a $Y$-sufficient representation whose pruning
yields $\mathrm{MB}^{\mathcal H}(Y)$.

\subsection{Identifiability Gains from Simulator Counterfactuals}
\label{sec:theory:ident}

Constraint-based discovery on observational records identifies at most the
observational equivalence class of the local causal graph, leaving some edge
orientations unresolved
\citep{spirtes2000causation,zhang2008causal,chickering2002optimal}.

\begin{proposition}[Resolving ambiguous edges via simulator counterfactuals]
\label{prop:counterfactual}
Simulator-internal counterfactual contrasts can resolve some CPDAG/PAG orientation
ambiguities by ruling out directions inconsistent with the induced effect,
making a subset of edges identifiable.
\end{proposition}

\subsection{Minimal Yet Sufficient Emergence Explanation via \texorpdfstring{$E_Y$}{E\_Y}}
\label{sec:theory:ees}

\textsc{CAMO} augments the computable Markov boundary
$\mathrm{MB}^{\mathcal H}(Y)$ with upstream causal pathways to form an
explanatory subgraph $E_Y$ (Eq.~\eqref{eq:ees_def}), enabling micro-to-macro
interpretation of emergence.

\begin{proposition}[Sufficiency and minimality of $E_Y$]
\label{prop:ees}
The subgraph $E_Y$ is sufficient for local prediction and intervention on $Y$,
since it contains $\mathrm{MB}^{\mathcal H}(Y)$.
Moreover, $E_Y$ is minimal for mechanistic explanation: it retains exactly the
causal structure required to connect each boundary variable to upstream
micro/meso-level root causes, and contains no superfluous nodes or edges.
\end{proposition}
\begin{table*}[t]
\centering
\caption{Comparison of different causal discovery methods for factor discovery on the O2O delivery simulation (Markov boundary vs.\ full-ancestor targets). All LLM-based baselines are evaluated using DeepSeek-V3.2.}
\label{tab:meituan_mb}
\small
\setlength{\tabcolsep}{4pt}
\renewcommand{\arraystretch}{1.}

\begin{tabular}{lccccccccc}
\toprule
Method &
MB$\uparrow$ & AN$\uparrow$ & OT$\downarrow$ &
MB-Prec$\uparrow$ & MB-Rec$\uparrow$ & MB-F1$\uparrow$ &
Anc-Prec$\uparrow$ & Anc-Rec$\uparrow$ & Anc-F1$\uparrow$ \\
\midrule
Single-round (Text-only)
 & 0 & \underline{8} & 6
 & 0.00 & 0.00 & 0.00
 & 0.20 & 0.41 & 0.27 \\
Single-round (Data-grounded)
 & \underline{1} & \underline{8} & 6
 & \underline{0.20} & \underline{0.50} & \underline{0.29}
 & 0.25 & \underline{0.61} & \underline{0.35} \\
Single-round (CoT)
 & 0 & \underline{8} & 5
 & 0.00 & 0.00 & 0.00
 & 0.26 & 0.41 & 0.32 \\
Multi-round (No causal feedback)
 & 0 & \underline{8} & \underline{4}
 & 0.00 & 0.00 & 0.00
 & \underline{0.27} & 0.41 & 0.33 \\
\midrule
\textsc{CAMO} (Ours)
 & \textbf{2} & \textbf{9} & \textbf{0}
 & \textbf{1.00} & \textbf{1.00} & \textbf{1.00}
 & \textbf{0.95} & \textbf{1.00} & \textbf{0.98} \\
\bottomrule
\end{tabular}
\end{table*}

\definecolor{groupgray}{gray}{0.93}
\begin{table*}[t]
\centering
\caption{Comparison of different causal discovery methods for causal structure recovery
on the O2O delivery simulation.
All LLM baselines use DeepSeek-V3.2; \textsc{CAMO} uses the LLM specified in parentheses.}
\label{tab:o2o_dag}
\small
\setlength{\tabcolsep}{3.8pt}
\renewcommand{\arraystretch}{1.}

\begin{tabular}{lccccccccccc}
\toprule
Method &
Prc$\uparrow$ & Rec$\uparrow$ & F1$\uparrow$ & Acc$\uparrow$ & Anc-F1$\uparrow$ &
FPR$\downarrow$ & SHD$\downarrow$ &
Add.$\downarrow$ & Mis.$\downarrow$ & Rev.$\downarrow$ & Unor.$\downarrow$ \\
\midrule

\rowcolor{groupgray}
\multicolumn{12}{l}{\emph{Statistical causal discovery (SCD)}} \\
PC
 & 0.20 & 0.38 & 0.26 & 0.61 & 0.42 & 0.34 & 23 & 15 & 3 & 5 & \textbf{0} \\
FCI
 & 0.13 & 0.23 & 0.17 & 0.58 & 0.19 & 0.34 & 29 & 19 & \textbf{0} & 6 & 4 \\
GES
 & 0.12 & 0.15 & 0.14 & 0.65 & 0.19 & 0.24 & 20 & 9 & 6 & 5 & \textbf{0} \\
MMHC
 & 0.12 & 0.15 & 0.13 & 0.64 & 0.12 & 0.25 & 20 & 9 & 5 & 6 & \textbf{0} \\

\rowcolor{groupgray}
\multicolumn{12}{l}{\emph{Pure LLM-based methods}} \\
Efficient-CDLMs
 & 0.71 & 0.77 & 0.74 & 0.90 & 0.91 & 0.07 & 7 & 4 & \underline{1} & \textbf{0} & 2 \\
MAC
 & 0.67 & 0.77 & 0.71 & 0.89 & 0.92 & 0.08 & 8 & 5 & 2 & \textbf{0} & \underline{1} \\
PAIRWISE
 & 0.37 & 0.54 & 0.44 & 0.75 & 0.46 & 0.20 & 15 & 9 & 5 & \underline{1} & \textbf{0} \\

\rowcolor{groupgray}
\multicolumn{12}{l}{\emph{Hybrid SCD+LLM methods}} \\
SCD (PC)-LLM
 & 0.42 & 0.77 & 0.54 & 0.76 & 0.60 & 0.24 & 18 & 15 & 3 & \textbf{0} & \textbf{0} \\
ReAct
 & 0.60 & 0.69 & 0.64 & 0.86 & 0.83 & 0.10 & 10 & 6 & 2 & \textbf{0} & 2 \\
LLM-KBCI
 & 0.85 & 0.85 & 0.85 & 0.94 & 0.74 & \underline{0.03} & 3 & \textbf{1} & \underline{1} & \underline{1} & \textbf{0} \\

\midrule
\textsc{CAMO} (Qwen3-235B-A22B)
 & 0.80 & \underline{0.92} & 0.86 & 0.94 & 0.97 & 0.05 & 4 & 3 & \underline{1} & \textbf{0} & \textbf{0} \\
\textsc{CAMO} (DeepSeek-R1-Qwen-32B)
 & 0.59 & \textbf{1.00} & 0.74 & 0.88 & 0.76 & 0.15 & 9 & 9 & \textbf{0} & \textbf{0} & \textbf{0} \\
\textsc{CAMO} (GPT-5 mini)
 & \underline{0.87} & \textbf{1.00} & \underline{0.93} & \underline{0.97} & \underline{0.97} & \underline{0.03} & \underline{2} & \underline{2} & \textbf{0} & \textbf{0} & \textbf{0} \\
\textsc{CAMO} (DeepSeek-V3.2)
 & \textbf{0.93} & \textbf{1.00} & \textbf{0.96} & \textbf{0.99} & \textbf{1.00} & \textbf{0.02} & \textbf{1} & \textbf{1} & \textbf{0} & \textbf{0} & \textbf{0} \\
\textsc{CAMO} (Gemma3-27B)
 & 0.71 & \underline{0.92} & 0.80 & 0.92 & 0.82 & 0.08 & 6 & 5 & \underline{1} & \textbf{0} & \textbf{0} \\
\bottomrule
\end{tabular}
\end{table*}

\begin{figure*}[t]
\centering
\includegraphics[width=\textwidth]{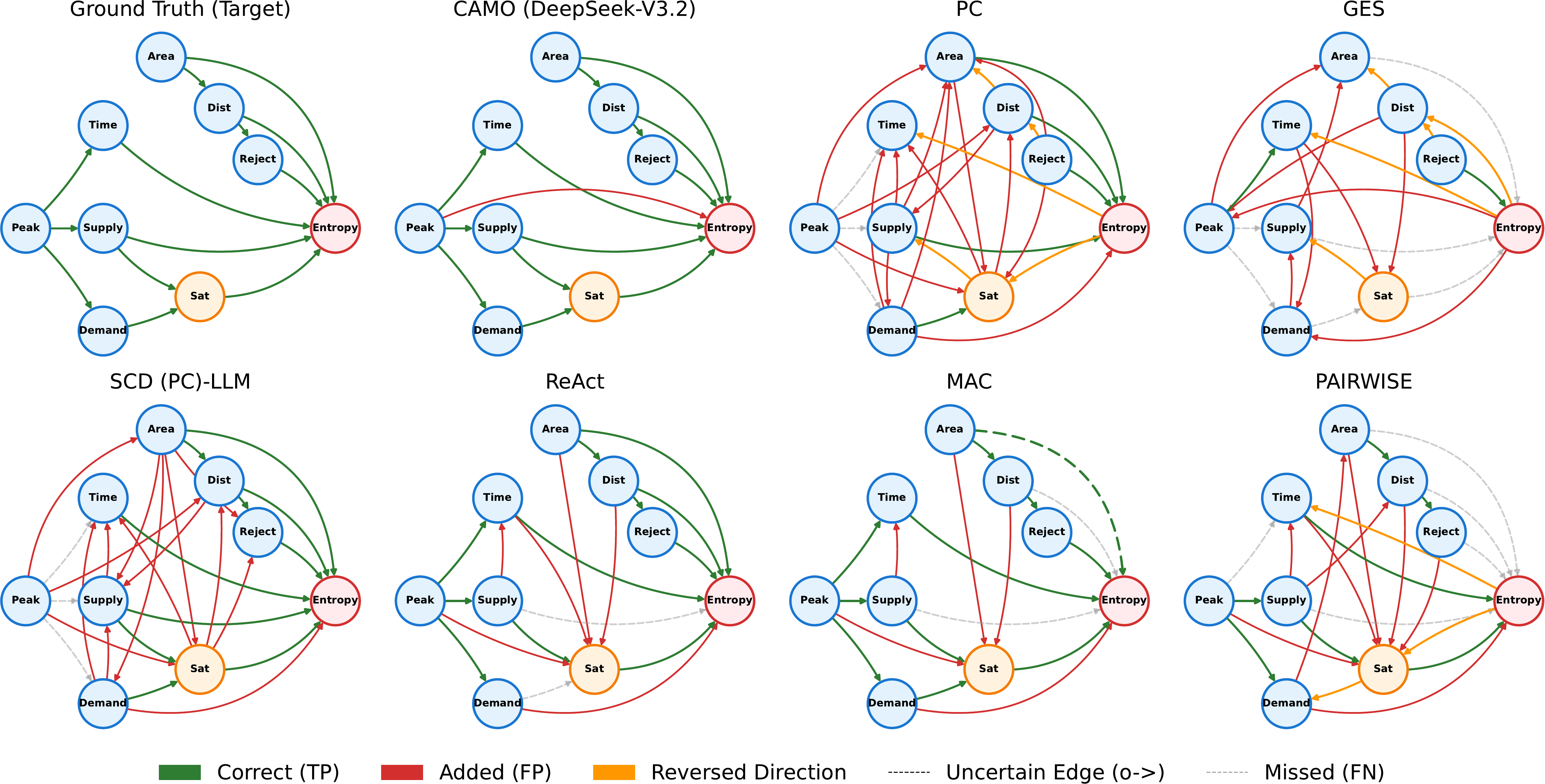}
\caption{Qualitative comparison of recovered causal structures (O2O delivery simulation; projected). Full visualizations for all methods are in Figure~\ref{app:causal_graphs_all_methods}.}
\label{fig:meituan_gt_vs_ours}
\end{figure*}

\begin{table*}[t]
\centering
\caption{Intervention ranking without ground-truth causal graphs (higher is better). All LLM-based methods use DeepSeek-V3.2. Results are averaged over 3 runs.}
\label{tab:interventional_utility}
\small
\vspace{-2mm}
\setlength{\tabcolsep}{2.5pt}
\renewcommand{\arraystretch}{0.95}
\begin{tabular}{lcccccccccc}
\toprule
& \multicolumn{3}{c}{Smallville (Coord.)}
& \multicolumn{3}{c}{AgentSociety (Polar.)}
& \multicolumn{3}{c}{AgentSociety (Inflam.)} \\
\cmidrule(lr){2-4}\cmidrule(lr){5-7}\cmidrule(lr){8-10}
Method
& Precision@5 & MAP@5 & MRR
& Precision@5 & MAP@5 & MRR
& Precision@5 & MAP@5 & MRR \\
\midrule

PC
& \underline{0.40} & 0.22 & 0.25
& 0.30 & 0.32 & \underline{0.75}
& 0.17 & 0.07 & 0.21 \\

MMHC
& 0.22 & 0.14 & 0.31
& 0.40 & 0.56 & \textbf{1.00}
& 0.16 & 0.08 & 0.22 \\

PAIRWISE
& \underline{0.40} & 0.22 & 0.25
& 0.30 & 0.32 & \underline{0.75}
& 0.18 & 0.09 & 0.24 \\

SCD (PC)+LLM
& 0.18 & 0.09 & 0.24
& \underline{0.50} & \underline{0.63} & \textbf{1.00}
& 0.16 & 0.08 & 0.22 \\

LLM-KBCI
& \underline{0.40} & \underline{0.28} & \underline{0.33}
& \underline{0.50} & 0.46 & 0.50
& \underline{0.26} & \underline{0.16} & \underline{0.27} \\

\midrule
\textsc{CAMO}(Ours)
& \textbf{0.60} & \textbf{0.64} & \textbf{0.50}
& \textbf{0.60} & \textbf{0.84} & \textbf{1.00}
& \textbf{0.60} & \textbf{0.71} & \textbf{0.61} \\

\bottomrule
\end{tabular}
\vspace{-2mm}
\end{table*}





\begin{figure}[t]
\centering
\includegraphics[width=\linewidth]{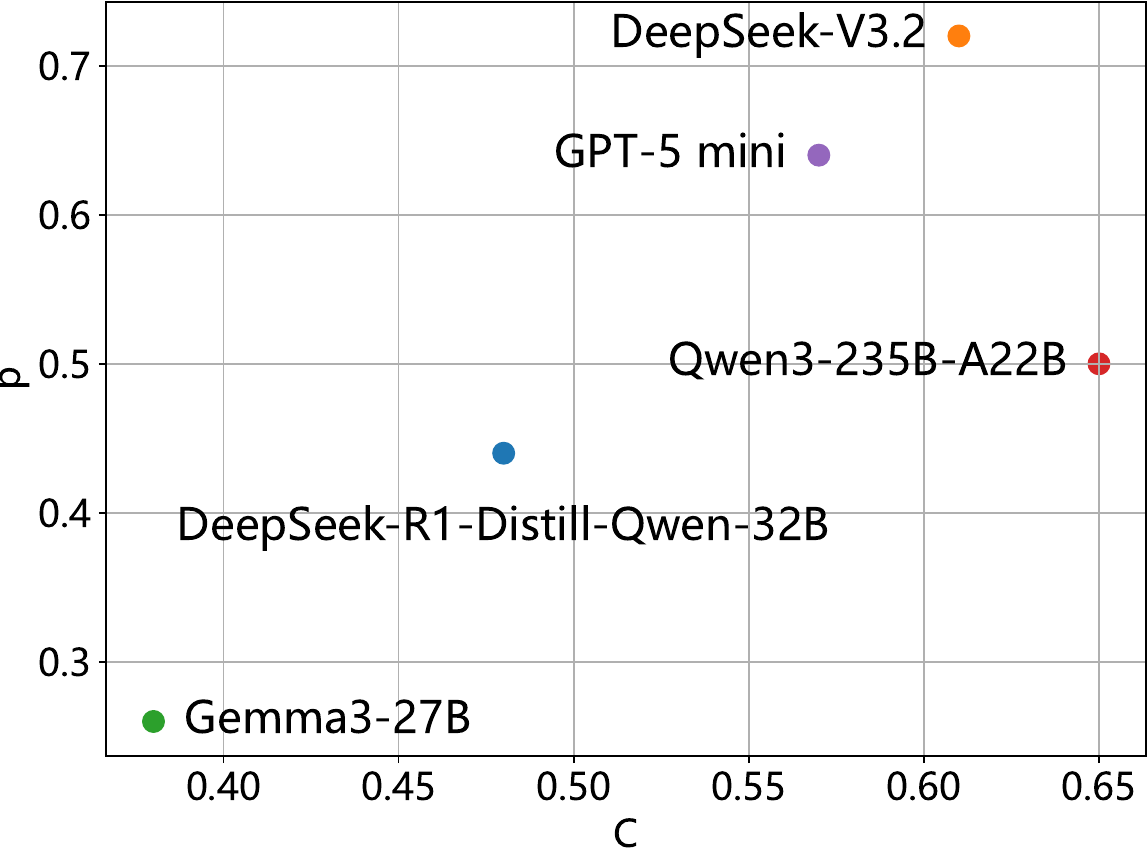}
\captionsetup{skip=2pt,font=small}
\vspace{-3mm}
\caption{Estimated $(p,C)$ across LLMs.}
\label{fig:pc_scatter}
\end{figure}

\begin{table}[t]
\centering
\captionsetup{skip=2pt,font=small}
\caption{Ablation on core components (O2O simulation); evaluated using GPT-5 mini.}
\label{tab:ablation}

\small
\setlength{\tabcolsep}{3pt}
\renewcommand{\arraystretch}{1.3}
\resizebox{\linewidth}{!}{%
\begin{tabular}{lcccccc}
\toprule
Variant &
MB-F1$\uparrow$ &
Anc-F1$\uparrow$ &
OT$\downarrow$ &
FPR$\downarrow$ &
SHD$\downarrow$ &
Unor.$\downarrow$ \\
\midrule
Full \textsc{CAMO}
 & \textbf{0.80} & \underline{0.90} & \underline{1} & \textbf{0.04} & \textbf{5} & \textbf{0} \\
\midrule
w/o A1\&A2
 & 0.00 & 0.46 & \textbf{0} & 0.08 & 23 & \textbf{0} \\
w/o A3-add
 & \underline{0.67} & 0.74 & \underline{1} & \textbf{0.04} & 9 & \textbf{0} \\
w/o A5 (no counterfactual)
 & \textbf{0.80} & 0.79 & 3 & \underline{0.05} & 10 & \textbf{0} \\
w/o competing hypotheses
 & \underline{0.67} & \textbf{0.91} & 2 & \textbf{0.04} & \underline{6} & \textbf{0} \\
\bottomrule
\end{tabular}
}
\end{table}

\section{Experiments}
\label{sec:experiments}

We evaluate \textsc{CAMO} under the following research questions:

\begin{itemize}
 \item \textbf{RQ1.}
Can \textsc{CAMO} recover the causal structure linking micro- and meso-level mechanisms to the macro outcome $Y$?

    \item \textbf{RQ2.}
Can the learned causal representation correctly identify effective interventions, namely actions on causes of $Y$ that reliably change $Y$?

\end{itemize}

\subsection{Factor Discovery and Causal Structure Recovery (RQ1)}
\label{sec:exp:meituan}

\paragraph{Simulation setup.}
We construct a benchmark consisting of an interactive LLM-empowered multi-agent simulation of an online-to-offline (O2O) delivery platform, built upon a \emph{minimal yet sufficient} ground-truth
causal structure calibrated to real-world statistics from Meituan Research~\citep{wang2025meituan}.
This ground-truth structure is used exclusively for evaluation and is never
exposed to \textsc{CAMO}.
We quantify emergence using the indicator of \citet{yu2025unlocking,yu2024beyond}; full simulation details see
Appendix~\ref{app:o2o_simulation}.


\paragraph{Baselines.}
\emph{Factor discovery.}
We compare \textsc{CAMO} against LLM-based factor discovery baselines, including
single-round text-only, data-grounded, and chain-of-thought prompting, as well as
multi-round prompting without causal feedback.
Details and prompts see
Appendices~\ref{app:factor_baselines} and~\ref{app:factor_prompts}.

\emph{Causal structure recovery.}
We compare \textsc{CAMO} against three classes of baselines.
\emph{Statistical causal discovery (SCD)} methods\footnote{We do \emph{not} perform time-series causal discovery: each independent rollout is treated as a single observational unit and mapped to run-level features, yielding a cross-sectional dataset on which PC/FCI are applied as static baselines.} include
PC~\cite{spirtes1991algorithm}, FCI~\cite{spirtes2000causation},
GES~\cite{chickering2002optimal}, and MMHC~\cite{tsamardinos2006max},
all applied directly to observed variables.
\emph{Pure LLM-based methods} include Efficient-CDLMs~\cite{jiralerspong2024efficient},
MAC~\cite{le2024multi}, and PAIRWISE~\cite{kiciman2023causal}, which rely on LLM
reasoning to infer causal relations without explicit statistical tests.
\emph{Hybrid SCD+LLM methods} combine statistical discovery with LLM-based graph
refinement, including SCD-LLM, ReAct~\cite{yao2022react}, and
LLM-KBCI~\cite{takayama2024integrating}.
Details are provided
in Appendix~\ref{app:structure_base}.


\paragraph{Metrics.}

\emph{(i) Factor discovery for the local interface and upstream explanation.}
We evaluate discovered factors by comparing them with ground truth, using the target’s Markov boundary $B(Y)$ and ancestor set $\mathrm{An}(Y)$ as references.
Discovered factors are categorized as Markov-boundary (MB), ancestor-but-non-boundary (AN), or off-target (OT), and we report category counts as well as Precision, Recall, and F1 for MB recovery.
To assess whether the learned local interface and retained upstream structure cover the true causal ancestry relevant to emergence, we additionally report ancestor-level accuracy (Anc-F1).
Formal definitions are provided in Appendix~\ref{app:metrics:factor}.

\emph{(ii) Causal structure recovery for emergence explanation.}
To compare \textsc{CAMO} with methods operating solely on logged observables, we
report standard structure metrics---Precision, F1, Ancestor F1, FPR, SHD, and
Added/Missed/Reversed/Unoriented edge counts
(Appendix~\ref{app:metrics:structure})---computed on the observed-variable
projection $\Pi(\cdot)$, which removes constructed factors and maps their effects
back to observed variables
(Appendix~\ref{app:metrics:projection}).
This projection enables a fair comparison with statistical causal discovery
methods, which typically operate on fixed observed variables and do not propose
new factors.
For completeness, we also report results without projection; see
Appendix~\ref{app:llm_comparison}.


\paragraph{Model.}
\textsc{CAMO} is evaluated with multiple LLM backbones:
Qwen3-235B-A22B~\cite{yang2025qwen3},
DeepSeek-R1-Distill-Qwen-32B~\cite{deepseekai2025deepseekr1incentivizingreasoningcapability},
DeepSeek-V3.2~\cite{deepseekai2025deepseekv32},
GPT-5 mini~\cite{openai_gpt5mini_modelpage_2025} and Gemma3-27B~\cite{gemma_2025}.

\paragraph{Results.}
\textbf{Factor discovery.}
Table~\ref{tab:meituan_mb} shows that \textsc{CAMO} best recovers the target’s Markov boundary and
its full ancestor set, validating (i) a correct local causal interface for predicting/intervening on
$Y$ and (ii) sufficient upstream coverage for emergence explanation; notably, it substantially outperforms all baselines in both MB-F1 and Anc-F1.

\textbf{Structure recovery and minimal connecting explanatory subgraph.}
Against causal-structure-recovery baselines, \textsc{CAMO} achieves the strongest recovery in both
metrics and qualitative comparisons (Table~\ref{tab:o2o_dag},
Figure~\ref{fig:meituan_gt_vs_ours}; after observed-variable projection). Crucially, this validates
our Minimal Connecting Explanatory Subgraph objective: \textsc{CAMO} is \emph{sufficient} in that it
best preserves the full upstream explanatory pathways (highest Anc-F1), and \emph{minimal} in that it
introduces the fewest redundant edges (lowest Added, SHD, and FPR), yielding the most faithful and
compact graph. 

\textbf{Analysis without projection.}
For completeness, we also report metrics (Table~\ref{tab:o2o_CAMO_llms}) and visualizations (Figure~\ref{fig:causal-graphs-without}) without graph projection, and compare how different LLM backbones affect \textsc{CAMO}; notably, DeepSeek-V3.2 and GPT-5 mini yield higher Prc/F1 and more compact graphs (lower FPR/SHD). See Appendix~\ref{app:llm_comparison} for details.


\textbf{Refinement effectiveness.}
We analyze the A2--A3 iterative add/prune refinement in Appendix~\ref{app:add_prune}
(Figure~\ref{fig:add_prune_dynamics}). Motivated by Theorem~\ref{thm:geom}, Figure~\ref{fig:pc_scatter} reports estimated $(p,C)$; see Appendix~\ref{app:pc_analysis} and Appendix~\ref{app:pc_estimation} for analysis and estimation details.

\textbf{Ablation Study.}
Table~\ref{tab:ablation} and Figure~\ref{fig:ablation-o2o} show that ablating any component reduces recovery, with the most pronounced degradation when removing A1\&A2; see Appendix~\ref{app:ablation} for details.

\subsection{Identifying Effective Interventions without Ground Truth (RQ2)}
\label{sec:exp:stanford}

We evaluate \textsc{CAMO} in settings without ground-truth causal graphs,
asking whether the learned structure yields \emph{actionable} guidance for
intervening on emergent outcomes.

\paragraph{Simulation setup.}
We consider two LLM-agent simulation environments with complex emergent dynamics:
\textbf{Smallville}~\cite{park2023generative} and
\textbf{AgentSociety}~\cite{piao2025agentsociety}.
In Smallville, the target outcome is \emph{agent coordination};
in AgentSociety, we study \emph{opinion polarization} and the
\emph{spread of inflammatory messages}.

\paragraph{Metrics.}
Without ground-truth causal graphs, we evaluate whether the learned graph $G$ yields actionable guidance by ranking effective interventions ahead of ineffective ones.
Following the root-cause-analysis-style evaluation in \citet{zheng2024lemma,shen2024exploring}, we run Random Walk with Restart from $Y$ and use each candidate target node's RWR score (stationary visit probability) as the ranking score, where higher values indicate stronger graph-mediated relevance to $Y$.
We report Precision@K, MAP@K ($K{=}5$), and MRR; detailed definitions are in Appendix~\ref{app:metrics:intervention}.

\paragraph{Results.}
Table~\ref{tab:interventional_utility} shows \textsc{CAMO} consistently achieves
the best Precision@5 and MAP@5 on Polar., Inflam., and Smallville, indicating
more accurate top-$K$ \emph{effectiveness judgments}. Some baselines attain high
MRR (e.g., Polar MMHC/SCD: $1.00$) but lower MAP@5, suggesting weaker ranking
among the remaining interventions. Baselines perform worst on Inflammatory
Spread; see Appendix~\ref{app:case_study} for a detailed case study.

\section{Conclusion}
\label{sec:conclusion}
\textsc{CAMO} is a multi-agent framework for explaining emergence in LLM-agent simulations. It learns a computable Markov boundary and a minimal connecting explanatory subgraph. Experiments show strong factor recovery, accurate structure reconstruction, and actionable intervention guidance.

\section{Limitations}
\label{sec:limitations}

This work is scoped to causal discovery in LLM-agent simulations. \textsc{CAMO}
depends on LLM-derived priors (and thus on pretraining, prompts, and available
context), and counterfactual refinement cannot guarantee completeness. As a
\emph{local causal interface} around $Y$, it may miss important confounders that
are neither logged in $X_{\mathrm{obs}}$ nor expressible in $\mathcal H$, which
can leave residual confounding. Results are validated only with respect to the
simulator’s logging/dynamics and may not transfer beyond it. The iterative loop
also incurs a budget--accuracy trade-off from repeated simulator queries, and
learned actionable levers could be misused or over-interpreted. We partially
mitigate these issues by grounding updates in observational constraints and
simulator counterfactual tests, evaluating in controlled simulation settings,
and restricting claims to simulator-scoped, evidence-backed interventions.

\section{Ethical considerations}
Our study focuses on causal discovery and explanation in LLM-empowered multi-agent simulations, including an online-to-offline (O2O) delivery platform, where simulator mechanisms are calibrated using real-world statistics. We do not conduct experiments on human subjects, and we do not use, store, or release any personally identifiable information: the calibration relies only on non-identifying, aggregate statistics rather than individual-level records, and all reported analyses are performed on synthetic simulation rollouts and logs. Since CAMO can surface actionable ``levers'' in simulated systems, there is a risk that such interventions could be over-interpreted or misused outside the simulator; we therefore restrict claims to simulator-scoped, evidence-backed findings and evaluate only in controlled simulation settings.

\section*{Acknowledgments}
This work has been supported in part by National Key Research and Development Program of China (No.2025YFE0216300), National Natural Science Foundation of China (No. 62472306, No. 62441221), Tianjin University's 2024 Special Project on Disciplinary Development (No. XKJS-2024-5-9), Tianjin University Talent Innovation Reward Program for Literature \& Science Graduate Student (C1-2022-010), and Henan Province Key Research and Development Program (No.251111210500), Tianjin University Independent Innovation Project (No.2025XJ3-0043).

\bibliography{custom}

\appendix
\clearpage
\startcontents[app]  
\addcontentsline{toc}{section}{Appendix Index}

\printcontents[app]{}{1}{\setcounter{tocdepth}{1}}

\section{Theory Proofs}
\label{app:proofs}

This appendix provides proofs for the theoretical results in
\textsection~\ref{sec:theory}. Throughout, we focus on properties that are directly
implied by the refinement procedure and do not assume global identifiability or
set-wise convergence unless explicitly stated.

\subsection*{Proof of Theorem~\ref{thm:monotone}}
\label{app:proof:monotone}

At refinement step $t{+}1$, the data-realized representation is first augmented
by admitting a (possibly empty) set of new computable factors
$\mathcal Z_{t+1}$:
\begin{equation}
\tilde V^{(t+1)} = V^{(t)} \cup \mathcal Z_{t+1},
\end{equation}
and then pruned by removing variables that are conditionally redundant for
predicting the target $Y$ given the remaining variables. Define the pruned set
\begin{equation}
\label{eq:Rt_def}
R_t :=
\Bigl\{\,
W \in \tilde V^{(t+1)}
\;:\;
Y \perp\!\!\!\perp W \mid \tilde V^{(t+1)} \setminus \{W\}
\Bigr\},
\end{equation}
and the updated representation
\begin{equation}
V^{(t+1)} = \tilde V^{(t+1)} \setminus R_t .
\end{equation}

Let $B_t$ denote the Markov boundary of $Y$ in $V^{(t)}$, and let
$m_t := |\mathcal Z_{t+1}|$. Consider the candidate set
\begin{equation}
C_{t+1} := (B_t \cup \mathcal Z_{t+1}) \setminus R_t .
\end{equation}
For any $W \in R_t$, by definition~\eqref{eq:Rt_def},
\begin{equation}
Y \perp\!\!\!\perp W \mid \tilde V^{(t+1)} \setminus \{W\},
\end{equation}
so removing $W$ does not reduce the conditional predictive sufficiency of the
remaining variables for $Y$. Hence, there exists a sufficient conditioning set
$B'_{t+1} \subseteq C_{t+1}$ such that
\begin{equation}
Y \perp\!\!\!\perp V^{(t+1)} \setminus B'_{t+1} \mid B'_{t+1}.
\end{equation}
By construction,
\begin{equation}
|B'_{t+1}| \le |B_t| + m_t - |B_t \cap R_t|.
\end{equation}
Since $B_{t+1}$ is the \emph{minimal} sufficient set for $Y$ in $V^{(t+1)}$, it
follows that
\begin{equation}
|B_{t+1}| \le |B'_{t+1}| \le |B_t| + m_t,
\end{equation}
which proves~\eqref{eq:monotone}. Moreover, if $B_t \cap R_t \neq \varnothing$,
then at least one previously boundary variable is pruned and the bound tightens
accordingly.
\qed

\subsection*{Proof of Theorem~\ref{thm:geom}}
\label{app:proof:geom}

Let
\begin{equation}
F_t := I(Y; X_{\mathrm{obs}} \mid V^{(t)})
\end{equation}
denote the residual conditional dependence between the target $Y$ and the full
observation space at refinement step $t$. Let $E_t$ denote the event that the
refinement step achieves the contraction condition in~\eqref{eq:contraction}. By
assumption,
\begin{equation}
\Pr(E_t) \ge p.
\end{equation}

By definition of the refinement capability,
\begin{equation}
F_{t+1} \le
\begin{cases}
(1-C)F_t, & \text{if } E_t \text{ occurs},\\
F_t, & \text{otherwise}.
\end{cases}
\end{equation}

Taking conditional expectation and using $\Pr(E_t) \ge p$, we obtain
\begin{equation}
\mathbb{E}[F_{t+1} \mid F_t]
\le (1-pC) F_t .
\end{equation}
Iterating this inequality yields
\begin{equation}
\mathbb{E}[F_t] \le (1-pC)^t F_0,
\end{equation}
which establishes geometric decay of the residual dependence. This result
implies convergence in the sense that the representation stabilizes with respect
to predictive and interventional sufficiency, without requiring set-wise
convergence or uniqueness of the limiting Markov boundary.
\qed

\subsection*{Proof of Proposition~\ref{prop:counterfactual}}
\label{app:proof:counterfactual}

Consider an edge between variables $X$ and $Z$ whose orientation is ambiguous
under the observationally identified CPDAG or PAG. Define the simulator-implied
interventional effect
\begin{equation}
\begin{aligned}
\delta(X \!\rightarrow\! Z)
\;:=\;
\sup_{x,x'}\;
D_{\mathrm{TV}}\Big(
& P(Z \mid \mathrm{do}(X{=}x)), \\
& P(Z \mid \mathrm{do}(X{=}x'))
\Big).
\end{aligned}
\end{equation}

If $\delta(X \!\rightarrow\! Z) > 0$, then intervening on $X$ induces a detectable
change in the distribution of $Z$.
Under the locality/modularity assumption (i.e., the intervention is approximately surgical in the simulator), any candidate
orientation in which $X$ is not a (possibly indirect) cause of $Z$ would imply
invariance of $Z$ under $\mathrm{do}(X)$, contradicting the observed
interventional effect.

By the law of large numbers, empirical estimates of the interventional
distributions obtained from repeated paired simulator rollouts converge almost
surely to their simulator-implied limits. Consequently, candidate orientations
incompatible with the observed counterfactual contrast are eliminated with
probability one as the number of rollouts $n \to \infty$.
\qed

\paragraph{Controlling for invariance in probing.}
Importantly, our counterfactual probing is an in-simulator operation and the resulting orientations should be interpreted as simulator-specific causal evidence. To make simulator counterfactual probing as close as possible to a localized (approximately surgical) intervention, we generate paired rollouts that (i) start from the same simulator state/trajectory prefix, (ii) use aligned randomness via shared seeds (common random numbers) for the environment and scheduling, and (iii) keep all non-target simulator components fixed (environment dynamics, scheduler, and agent policy/system prompt), while only clamping the target variable/mechanism $X$.

\section{Supplementary Experimental Results}
\label{app:additional_results}

\begin{figure*}[!t]
\centering
\captionsetup{font=small}

\begin{subfigure}[t]{0.49\textwidth}
  \centering
  \includegraphics[width=\linewidth]{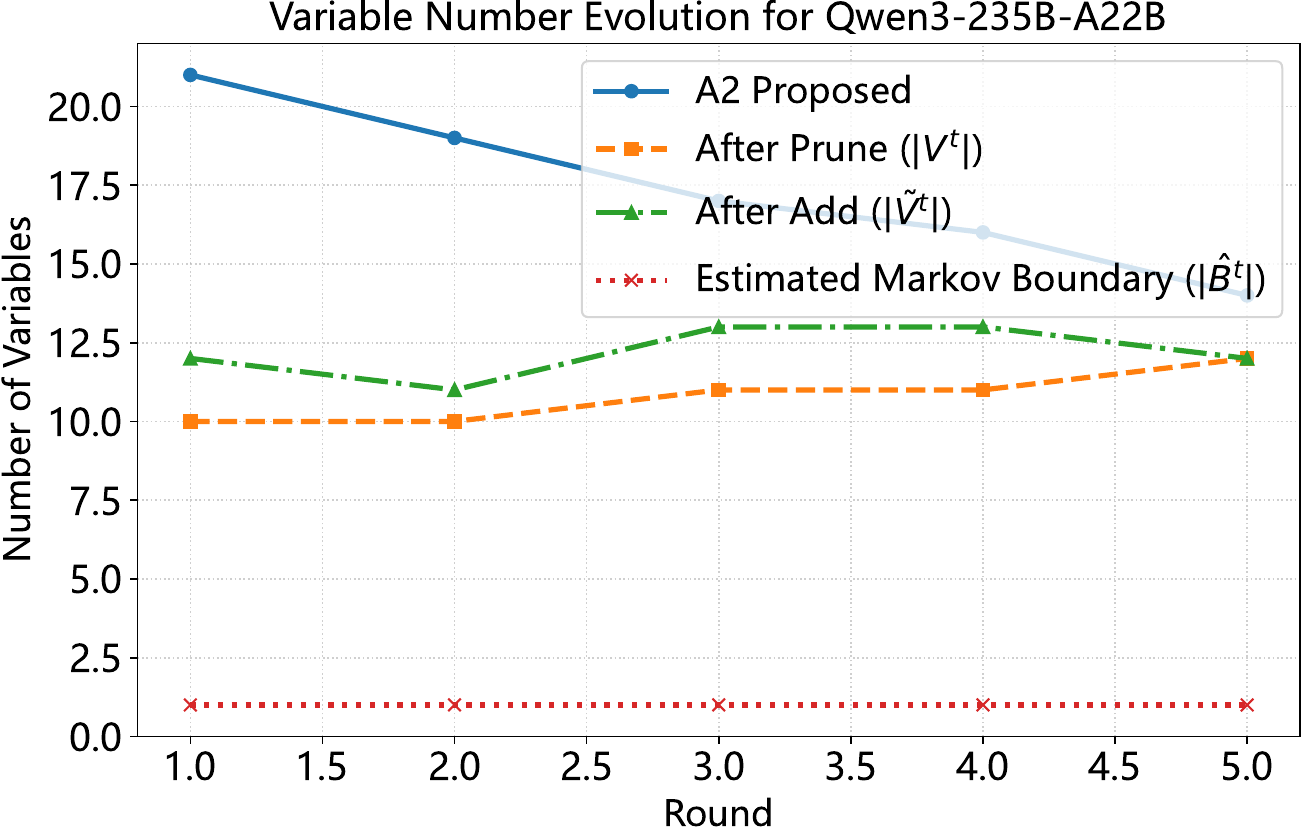}
  \caption{Qwen3-235B-A22B}
\end{subfigure}\hfill
\begin{subfigure}[t]{0.49\textwidth}
  \centering
  \includegraphics[width=\linewidth]{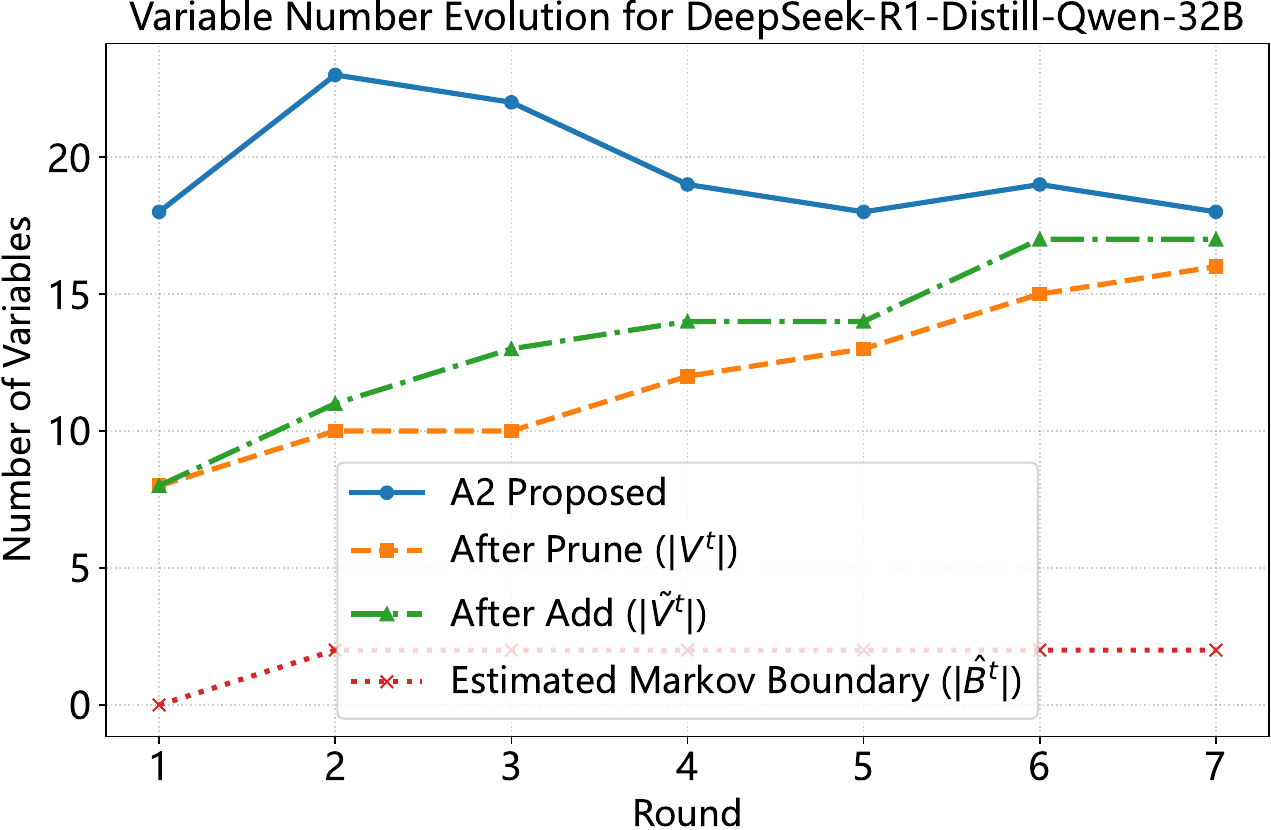}
  \caption{DeepSeek-R1-Distill-Qwen-32B}
\end{subfigure}

\vspace{2mm}

\begin{subfigure}[t]{0.49\textwidth}
  \centering
  \includegraphics[width=\linewidth]{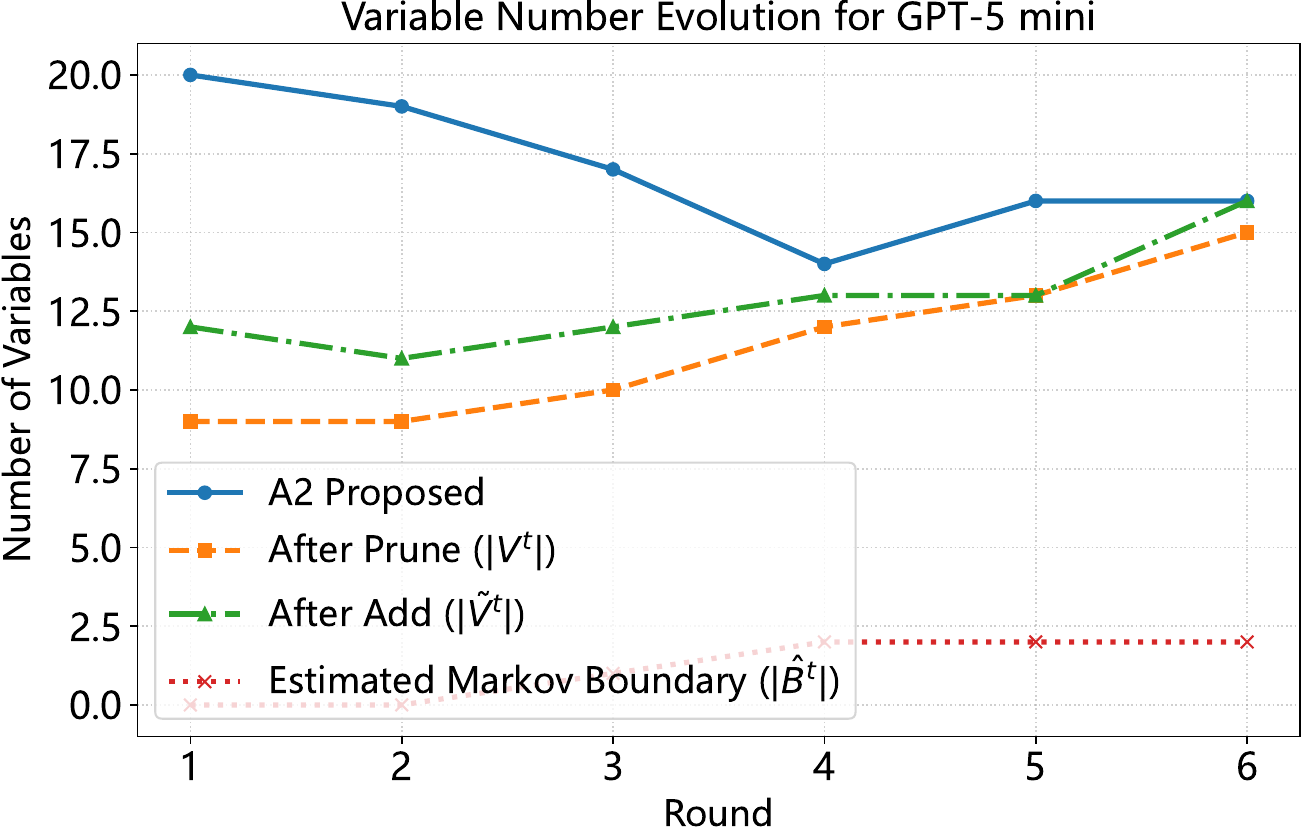}
  \caption{GPT-5 mini}
\end{subfigure}\hfill
\begin{subfigure}[t]{0.49\textwidth}
  \centering
  \includegraphics[width=\linewidth]{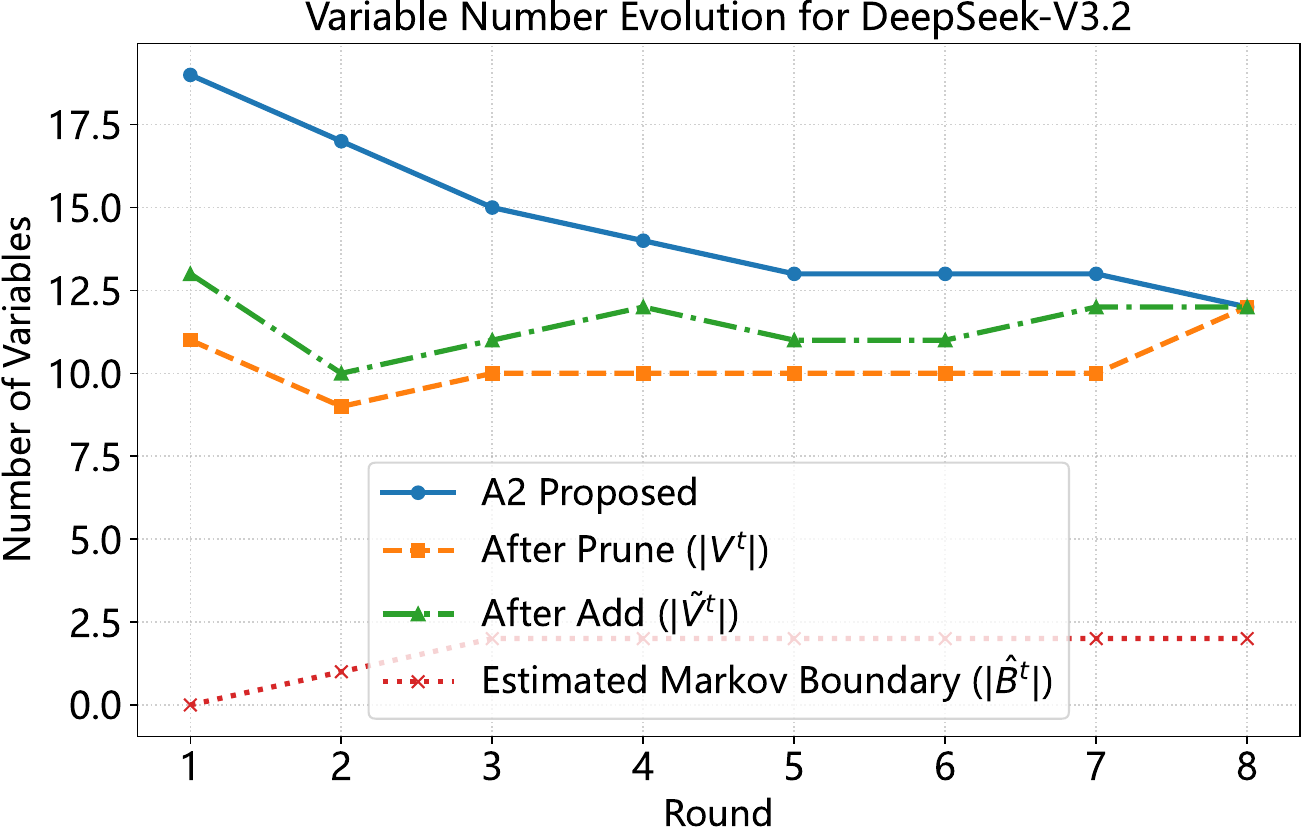}
  \caption{DeepSeek-V3.2}
\end{subfigure}

\vspace{2mm}


\begin{subfigure}[t]{0.49\textwidth}
  \centering
  \includegraphics[width=\linewidth]{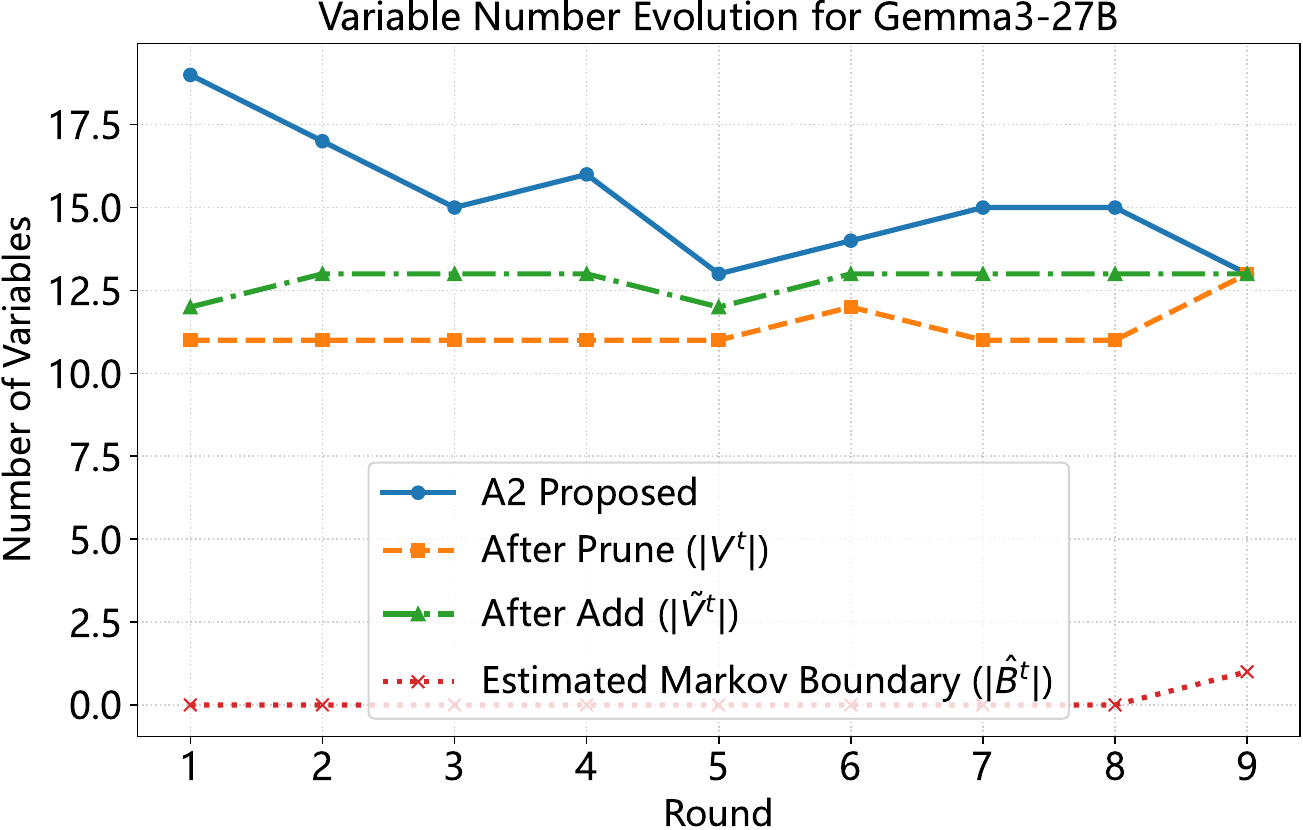}
  \caption{Gemma3-27B}
\end{subfigure}
\hfill

\caption{Add/prune dynamics of \textsc{CAMO} (A2--A3) under different LLM backbones.
We plot, per round, the A2 proposal size, sizes after prune/add, and the estimated Markov boundary (MB) size.
Pruning sharply compresses the proposal set, while add makes small corrections; meanwhile, the estimated MB progressively
recovers the ground-truth MB size (2 in this setting).}
\label{fig:add_prune_dynamics}
\end{figure*}

\subsection{A2--A3 Iterative Refinement (Add/Prune)}
\label{app:add_prune}

Figure~\ref{fig:add_prune_dynamics} illustrates how \textsc{CAMO} iteratively refines the candidate factor set via the A2--A3 loop.
Overall, the add/prune mechanism rapidly compresses an initially over-complete proposal into a compact, stable set, while
progressively recovering the ground-truth local neighborhood (Markov boundary size $=2$) across rounds.

\paragraph{(1) Candidate-set dynamics.}
At each round, A2 proposes a (potentially over-complete) set of candidate factors, which is then sharply reduced by A3's prune step.
After a few rounds, the candidate set size stabilizes at a much smaller scale, indicating effective regularization against uncontrolled growth.

\paragraph{(2) Role separation: prune vs.\ add.}
Across LLM backbones, a consistent pattern emerges: the prune step accounts for the dominant reduction (removing irrelevant factors),
whereas the add step introduces only modest corrections by re-inserting a small number of missed-but-necessary factors. This division of
labor prevents early over-pruning and improves recall without sacrificing compactness.

\paragraph{(3) Markov-boundary recovery.}
We additionally track the estimated Markov boundary size, which remains substantially smaller than the full candidate set throughout.
More importantly, it gradually approaches the ground-truth Markov boundary (size $2$ in this setting) and converges toward it across
rounds, demonstrating that the refinement loop increasingly identifies the true local causal neighborhood.

Together, these results suggest that the A2--A3 loop serves as a principled ``compress-and-correct'' procedure: it regularizes the search
space by pruning spurious variables, while preserving recall through selective additions, thereby progressively isolating the true local
causal structure.

\subsection{Effect of LLM backbones in the full factor space (no projection).}
\label{app:llm_comparison}

\begin{figure*}[!t]
    \centering

    \begin{subfigure}[t]{0.49\linewidth}
        \centering
 \includegraphics[width=\linewidth]{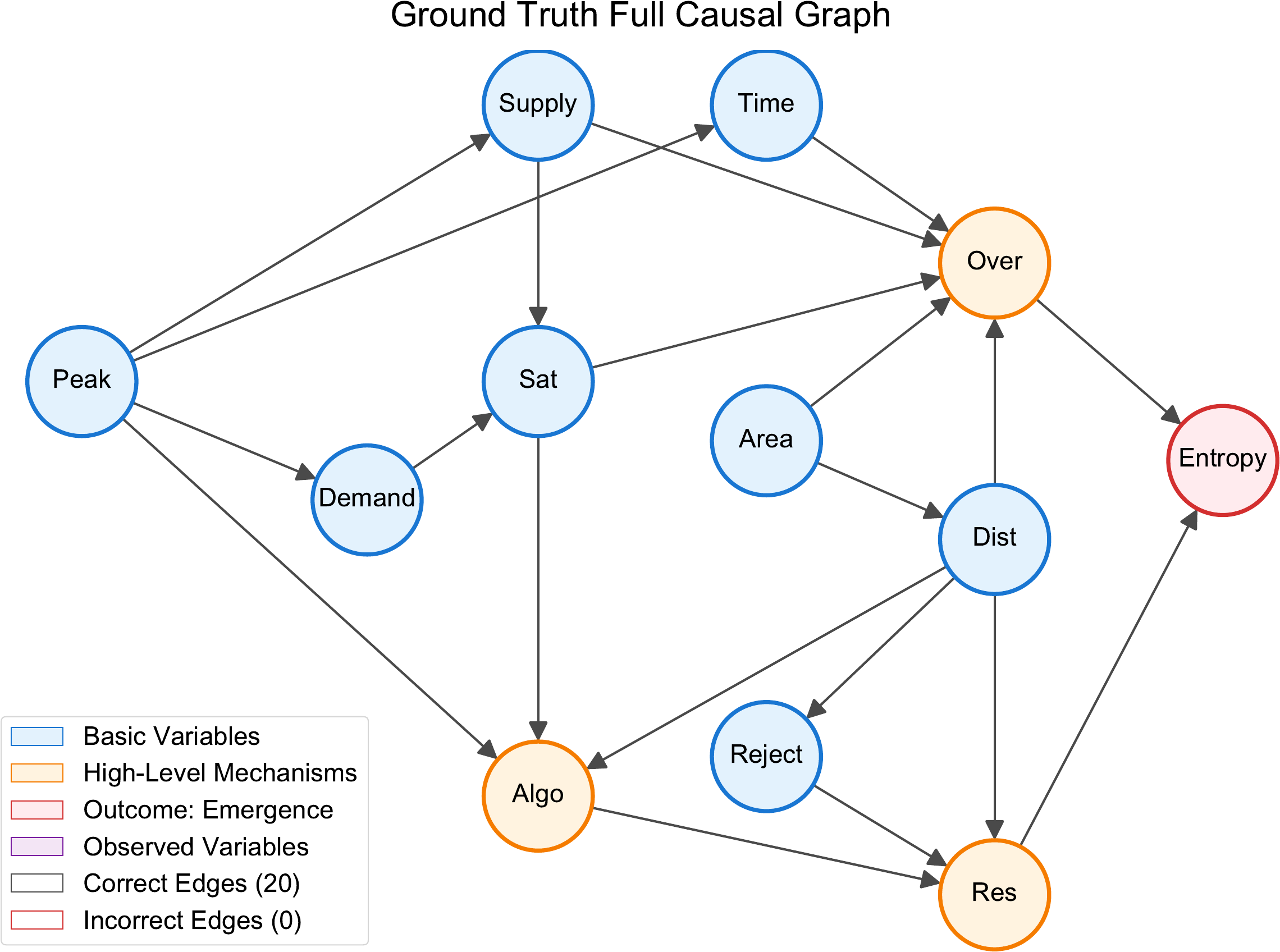}
        \caption{Ground Truth}
        \label{fig:cg-gt}
    \end{subfigure}\hfill
    \begin{subfigure}[t]{0.49\linewidth}
        \centering       
        \includegraphics[width=\linewidth]{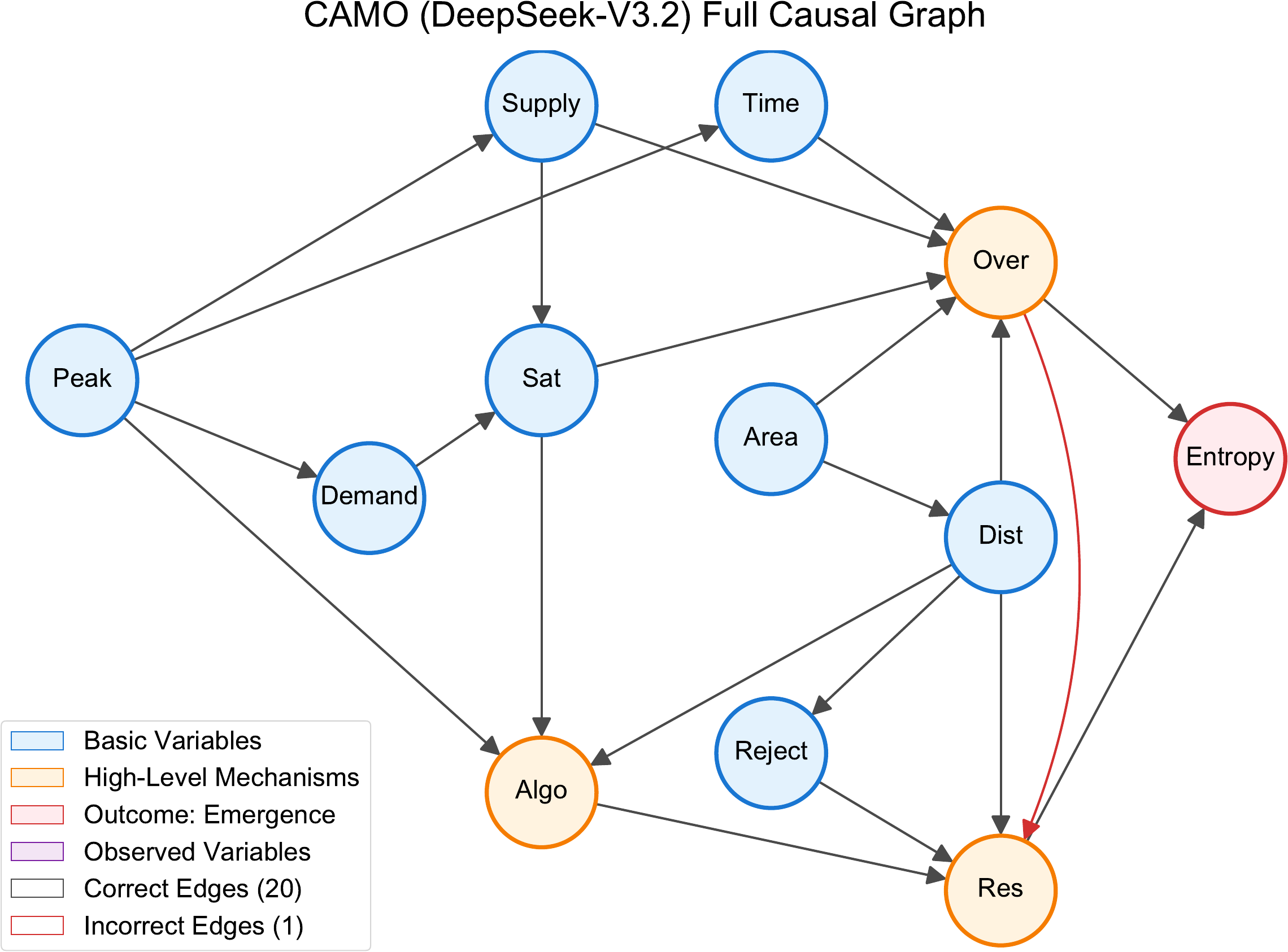}
        \caption{DeepSeek-V3.2}
        \label{fig:cg-dsv32}
    \end{subfigure}

    \vspace{0.35em}

    \begin{subfigure}[t]{0.49\linewidth}
        \centering
        \includegraphics[width=\linewidth]{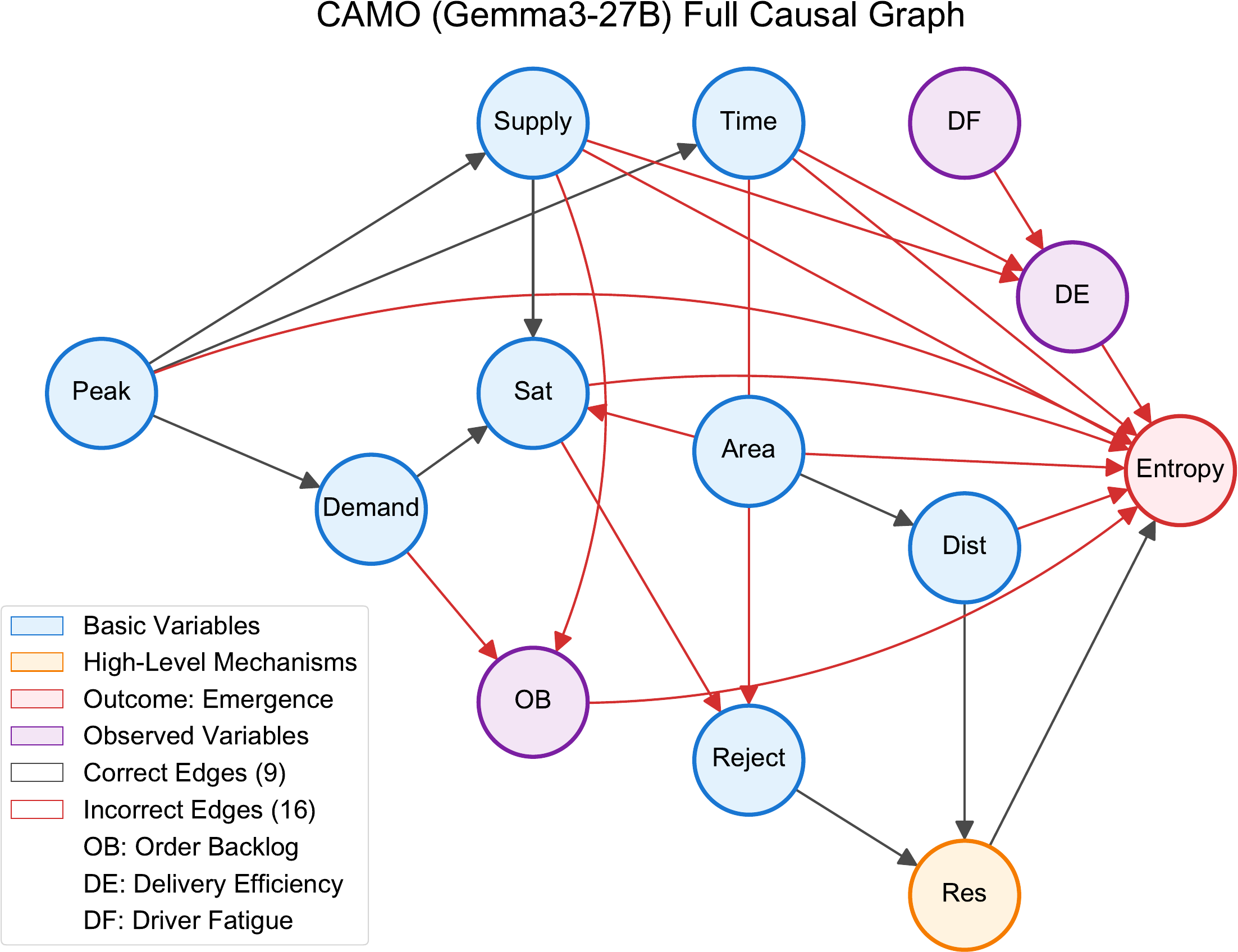}
        \caption{Gemma3-27B}
        \label{fig:cg-gemma}
    \end{subfigure}\hfill
    \begin{subfigure}[t]{0.49\linewidth}
        \centering
        \includegraphics[width=\linewidth]{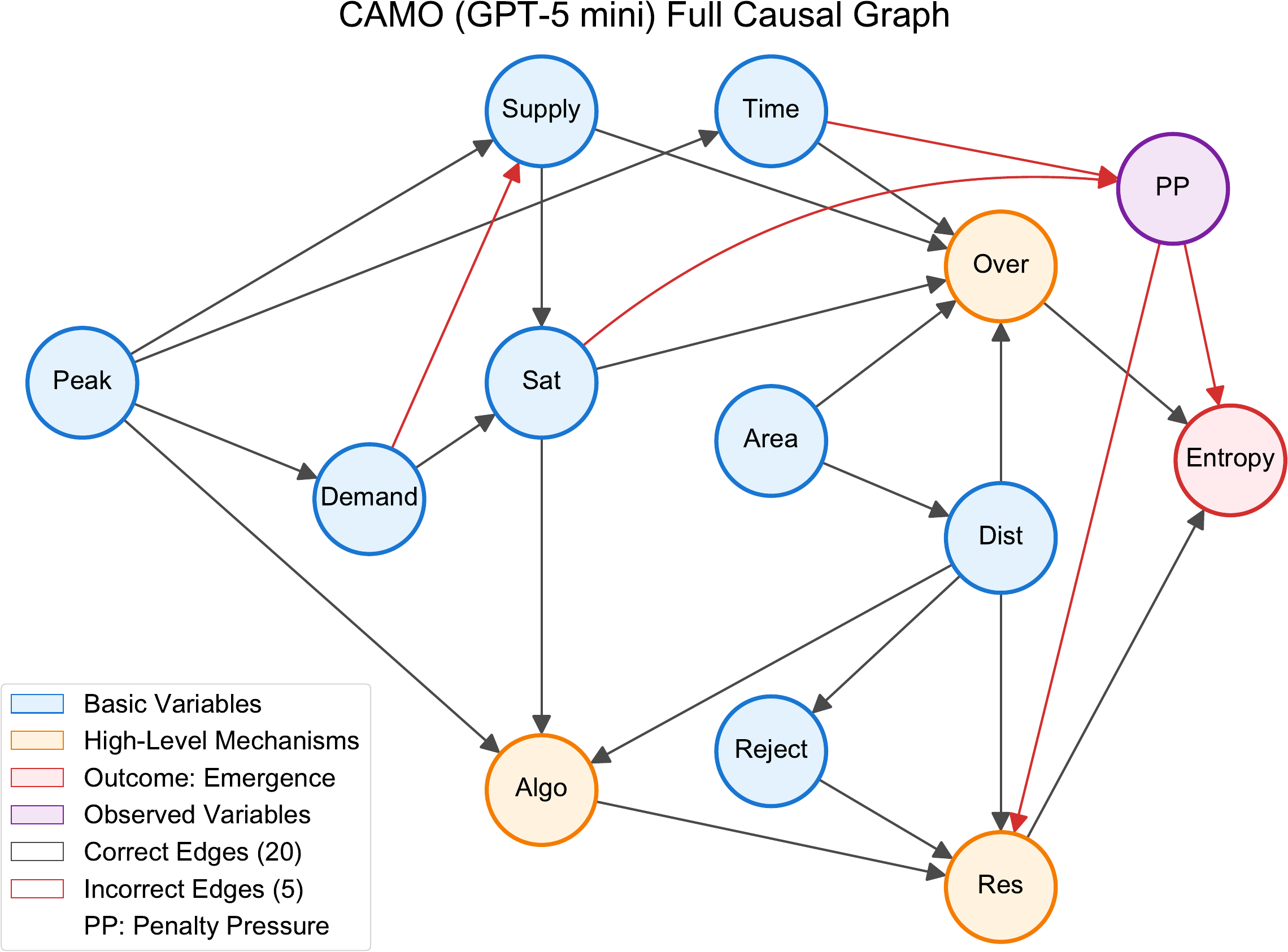}
        \caption{GPT-5 mini}
        \label{fig:cg-gpt5mini}
    \end{subfigure}

    \vspace{0.35em}

    \begin{subfigure}[t]{0.49\linewidth}
        \centering
        \includegraphics[width=\linewidth]{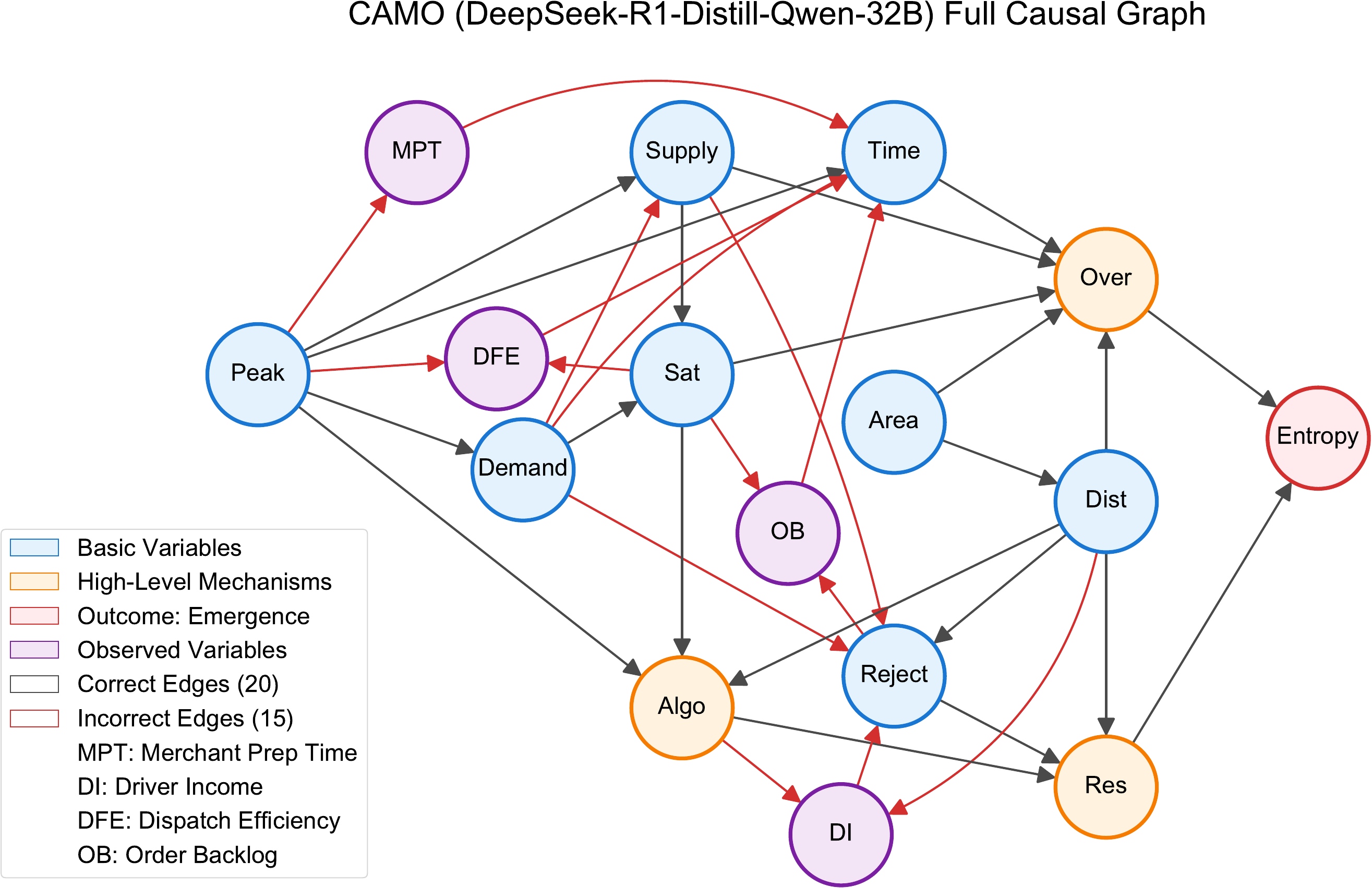}
        \caption{DeepSeek-R1-Distill-Qwen-32B}
        \label{fig:cg-ds32b}
    \end{subfigure}\hfill
    \begin{subfigure}[t]{0.49\linewidth}
        \centering
        \includegraphics[width=\linewidth]{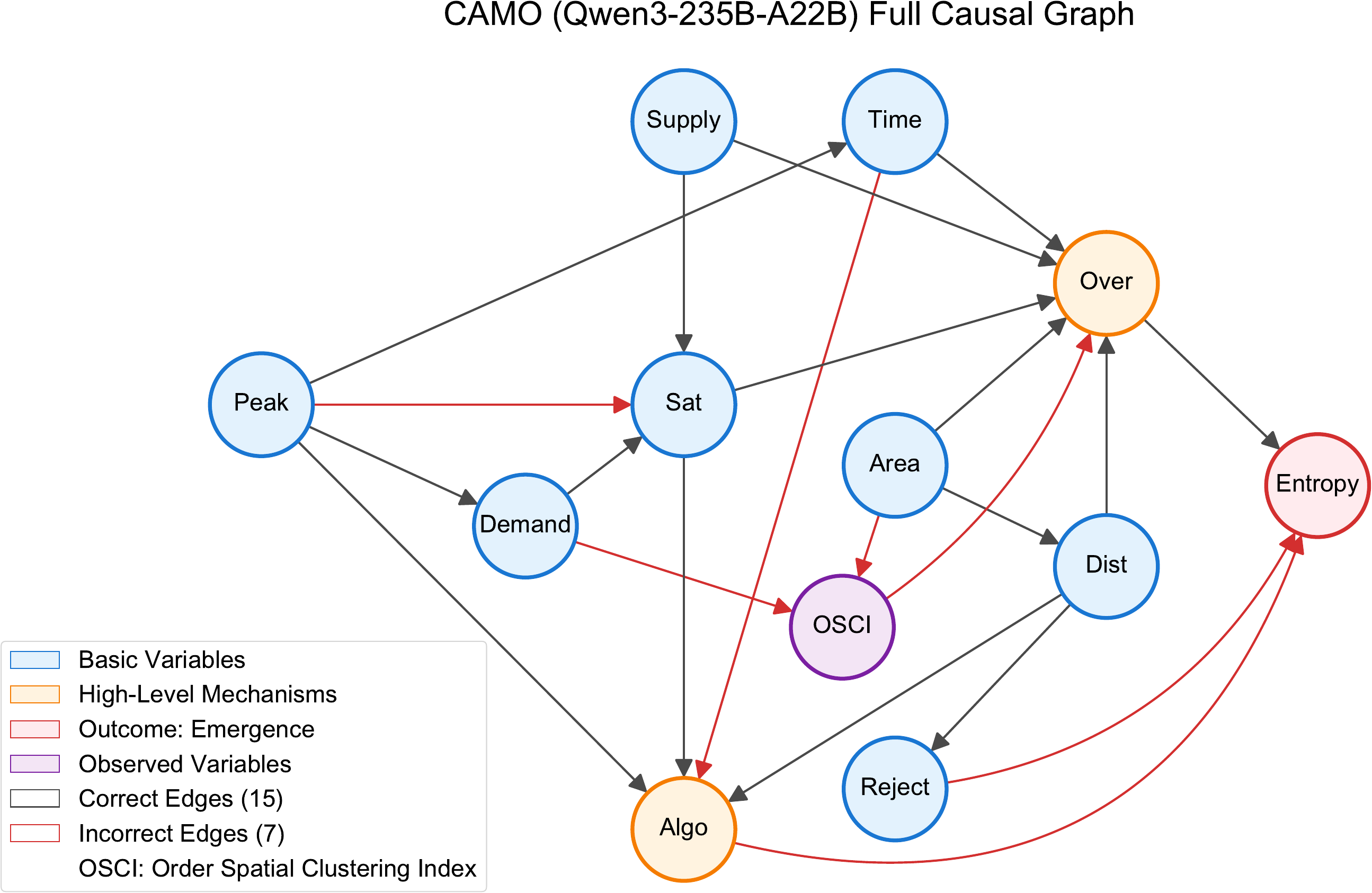}
        \caption{Qwen3-235B-A22B}
        \label{fig:cg-qwen}
    \end{subfigure}

    \caption{Qualitative comparison of recovered causal structures (O2O delivery simulation; without projection).}
    \label{fig:causal-graphs-without}
\end{figure*}

Table~\ref{tab:o2o_CAMO_llms} compares \textsc{CAMO} under different LLM backbones when evaluating
directly in the full induced factor space (without projection); qualitative graphs are shown in
Figure~\ref{fig:causal-graphs-without}.

\textbf{Sufficiency.}
Across backbones, \textsc{CAMO} maintains strong upstream coverage, attaining high recall for all
but Gemma3-27B. This suggests that once factors are elicited, the framework largely preserves the
target-relevant causal pathways.

\textbf{Minimality.}
Stronger backbones (DeepSeek-V3.2, GPT-5 mini) produce markedly more compact and faithful graphs,
achieving higher Prc/F1 and lower FPR/SHD, i.e., fewer redundant relations in the induced factor
space. Even with smaller backbones, \textsc{CAMO} still yields usable graphs and retains competitive
structure recovery, though with a looser sparsity--accuracy trade-off.

\begin{figure*}[!t]
\centering
\includegraphics[width=\textwidth]{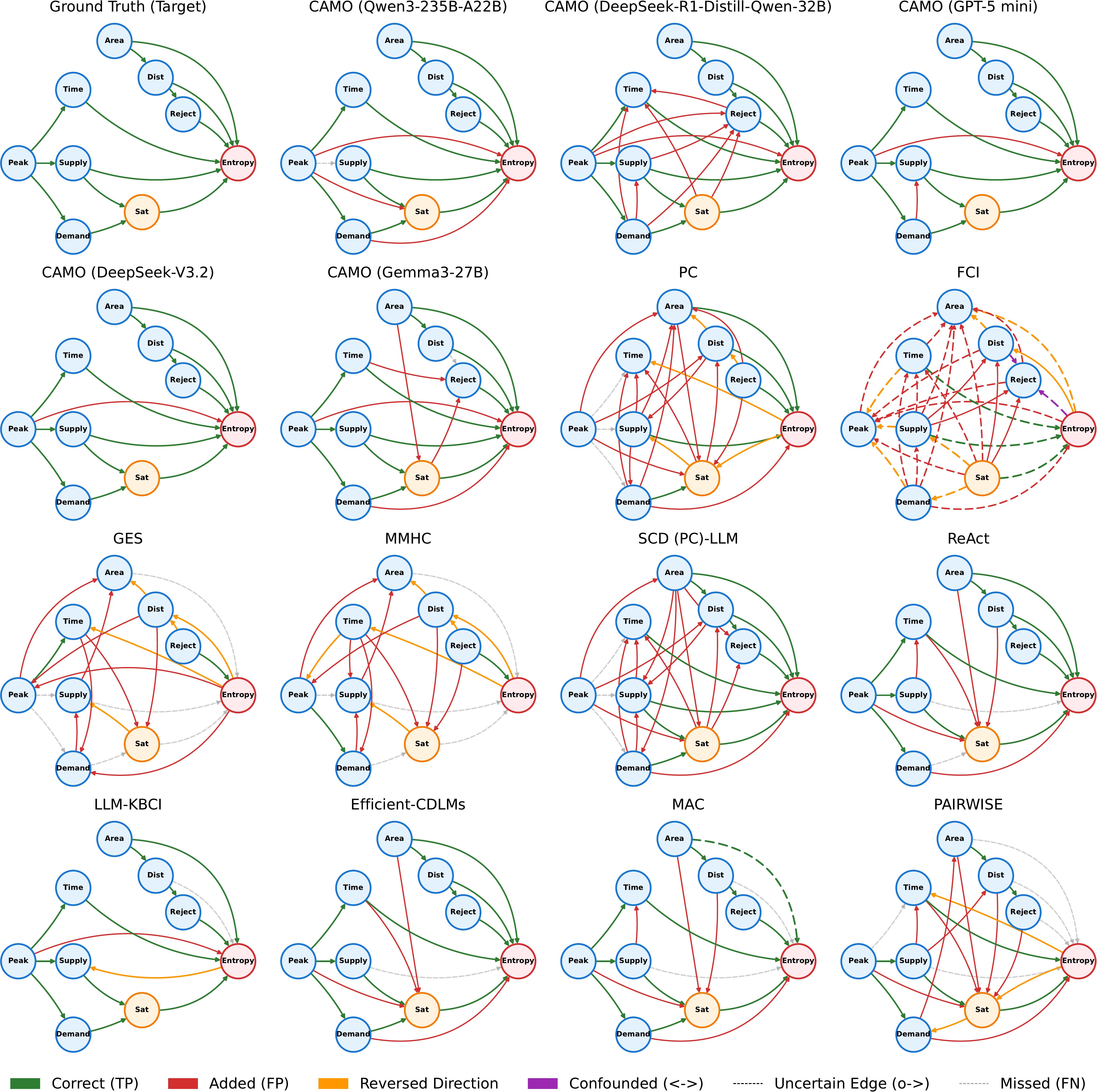}
\vspace{-2mm}
\caption{Recovered causal structures for \emph{all} methods on the O2O delivery simulation after applying the observed-variable projection. The figure provides a complete qualitative comparison complementing Figure~\ref{fig:meituan_gt_vs_ours}.}
\label{app:causal_graphs_all_methods}
\end{figure*}

\begin{table*}[t]
\centering
\caption{Causal structure recovery of CAMO on the O2O delivery simulation under different LLM backbones (results \emph{without graph projection}).}
\label{tab:o2o_CAMO_llms}
\small
\setlength{\tabcolsep}{3.8pt}
\renewcommand{\arraystretch}{1.08}

\begin{tabular}{lcccccccccccc}
\toprule
LLM Backbone &
Prc$\uparrow$ & Rec$\uparrow$ & F1$\uparrow$ & Acc$\uparrow$ & Anc-F1$\uparrow$ &
FPR$\downarrow$ & SHD$\downarrow$ &
Add.$\downarrow$ & Mis.$\downarrow$ & Rev.$\downarrow$ & Unor.$\downarrow$ \\
\midrule

Qwen3-235B-A22B &
0.68 & \underline{0.75} & \underline{0.71} & \underline{0.92} & \underline{0.79} &
\underline{0.05} & \underline{12} & \underline{7} & \underline{5} & \textbf{0} & \textbf{0} \\

DeepSeek-R1-Distill-Qwen-32B &
\underline{0.57} & \textbf{1.00} & 0.73 & \underline{0.94} & 0.62 &
0.07 & \underline{15} & \underline{15} & \textbf{0} & \textbf{0} & \textbf{0} \\

GPT-5 mini &
0.80 & \textbf{1.00} & 0.89 & 0.97 & 0.90 &
0.04 & 5 & 5 & \textbf{0} & \textbf{0} & \textbf{0} \\

DeepSeek-V3.2 &
\textbf{0.95} & \textbf{1.00} & \textbf{0.98} & \textbf{0.99} & \textbf{0.98} &
\textbf{0.01} & \textbf{1} & \textbf{1} & \textbf{0} & \textbf{0} & \textbf{0} \\

Gemma3-27B &
0.36 & 0.45 & 0.40 & 0.87 & 0.59 &
0.08 & 27 & 16 & 11 & \textbf{0} & \textbf{0} \\
\bottomrule
\end{tabular}
\end{table*}

\subsection{Case Study}
\label{app:case_study}

\begin{figure*}[!t]
    \centering
    \includegraphics[width=\textwidth]{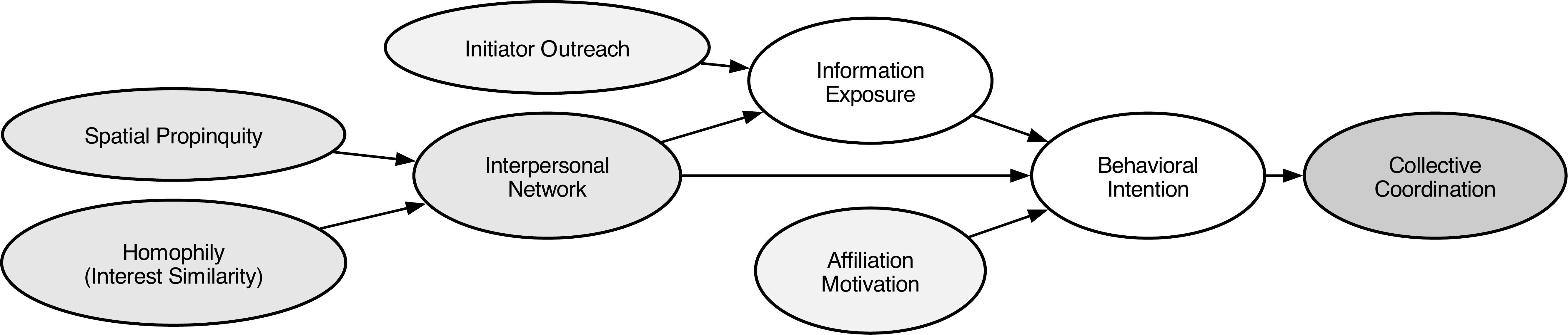}
    \caption{\textbf{Learned causal graph for agent coordination.}
    Visualization of a representative causal graph learned by CAMO under the
    \emph{agent coordination} emergent phenomenon.}
    \label{fig:causal_graph_coordination}
\end{figure*}

\begin{figure*}[!t]
    \centering
    \includegraphics[width=\textwidth]{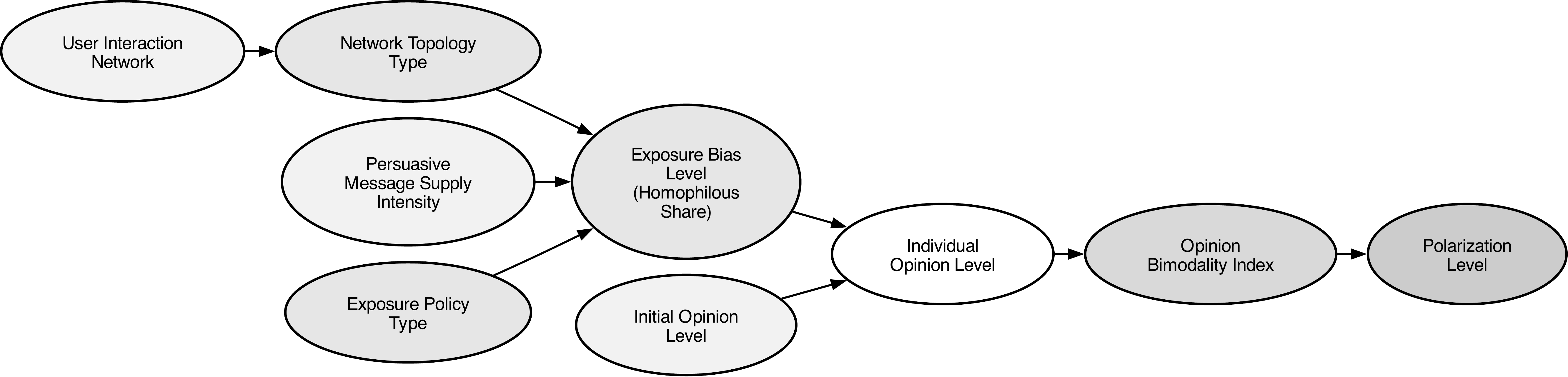}
    \caption{\textbf{Learned causal graph for opinion polarization.}
    Visualization of a representative causal graph learned by CAMO under the
    \emph{opinion polarization} emergent phenomenon.}
    \label{fig:causal_graph_polarization}
\end{figure*}

\begin{figure*}[!t]
    \centering
    \includegraphics[width=\textwidth]{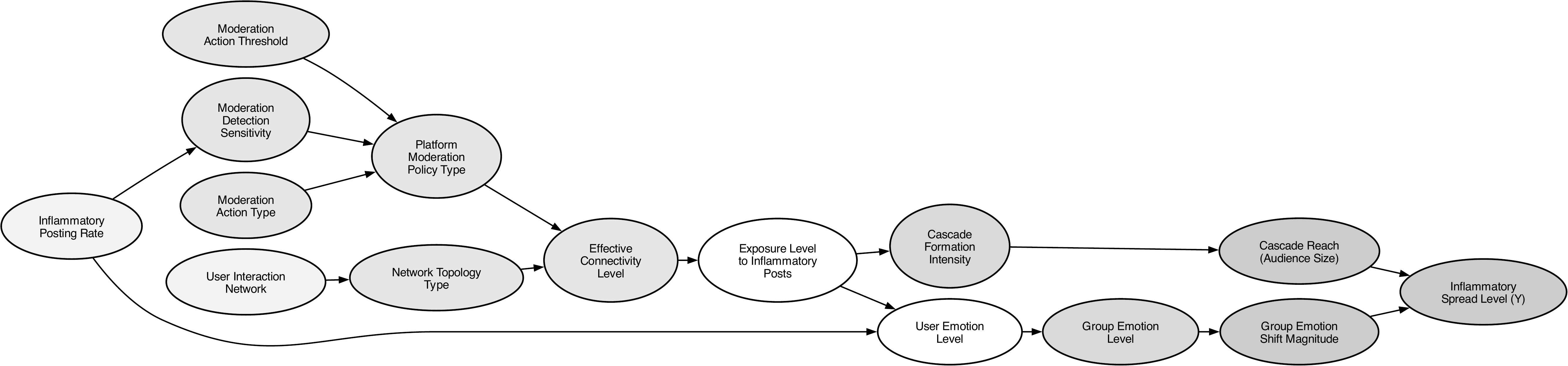}
    \caption{\textbf{Learned causal graph for inflammatory message spread.}
    Visualization of a representative causal graph learned by CAMO under the
    \emph{spread of inflammatory messages} emergent phenomenon.}
    \label{fig:causal_graph_inflammatory}
\end{figure*}

We present qualitative case studies of the causal graphs learned by \textsc{CAMO}
across multiple scenarios and emergent causal patterns. In particular, we
visualize the learned causal structures for three representative emergent
phenomena: (i) agent coordination (Figure~\ref{fig:causal_graph_coordination}),
(ii) opinion polarization (Figure~\ref{fig:causal_graph_polarization}), and
(iii) the spread of inflammatory messages
(Figure~\ref{fig:causal_graph_inflammatory}).

These examples illustrate how \textsc{CAMO} abstracts over low-level agent
interactions and recovers compact, high-level causal structures that capture
the key mechanisms underlying each emergent phenomenon. Importantly, all graphs
shown here are learned without access to any ground-truth causal structure.
They are therefore intended to support qualitative inspection and
interpretability, rather than serving as a quantitative evaluation benchmark.

For each setting, the learned graph highlights a small set of upstream factors
that are most actionable for intervention, while the remaining nodes represent
environmental conditions, intermediate system states, or emergent outcome
indicators. 

\begin{figure*}[t]
  \centering
  \includegraphics[width=\textwidth]{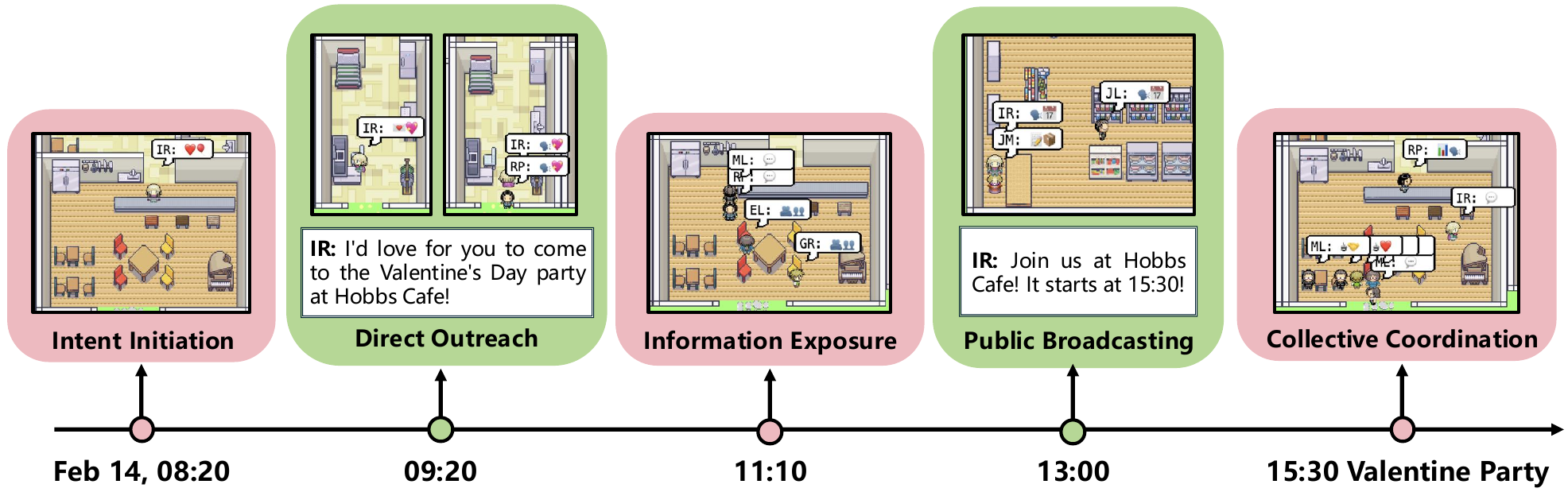}
  \caption{Timeline of an intervention trial in the agent coordination setting.
  The figure illustrates the emergence of collective coordination over time,
  progressing from intent initiation and direct outreach to information exposure
  and public broadcasting, culminating in collective coordination at the event
  time (15:30).}
  \label{fig:coordination_timeline}
\end{figure*}

In addition, for one intervention trial in the agent coordination setting, we
visualize the temporal evolution of the emergent coordination behavior based on
the learned causal graph (see Figure~\ref{fig:coordination_timeline}). This
timeline links the onset of the coordination phenomenon to the key causal
factors identified by \textsc{CAMO}, providing an intuitive view of how
interventions propagate through intermediate states and ultimately shape the
emergent outcome.

\begin{figure*}[!t]
\centering

\begin{subfigure}[t]{0.48\textwidth}
  \centering
  \includegraphics[width=\linewidth]{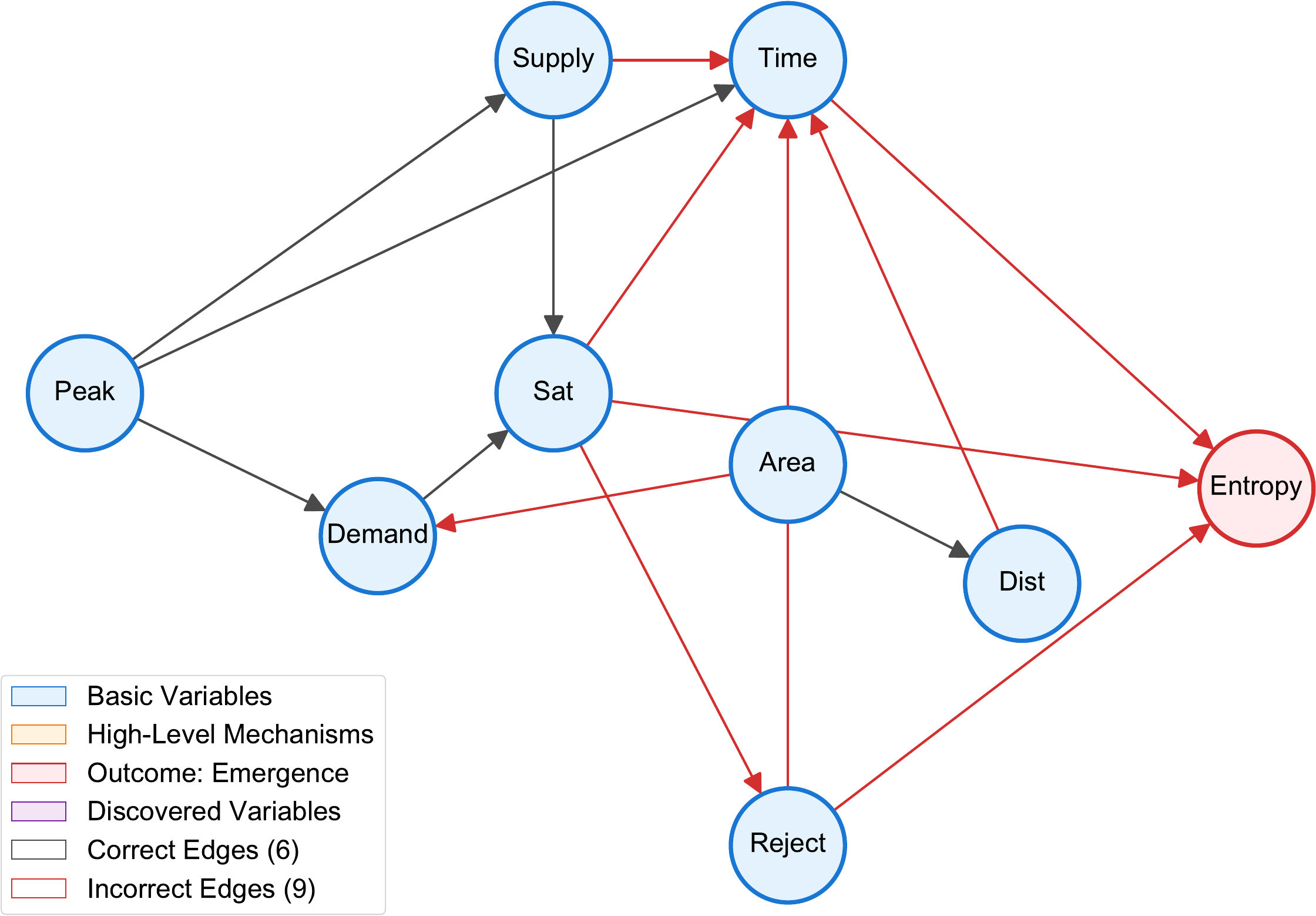}
  \caption{w/o A1\&A2}
  \label{fig:ablation-exp1}
\end{subfigure}\hfill
\begin{subfigure}[t]{0.48\textwidth}
  \centering
  \includegraphics[width=\linewidth]{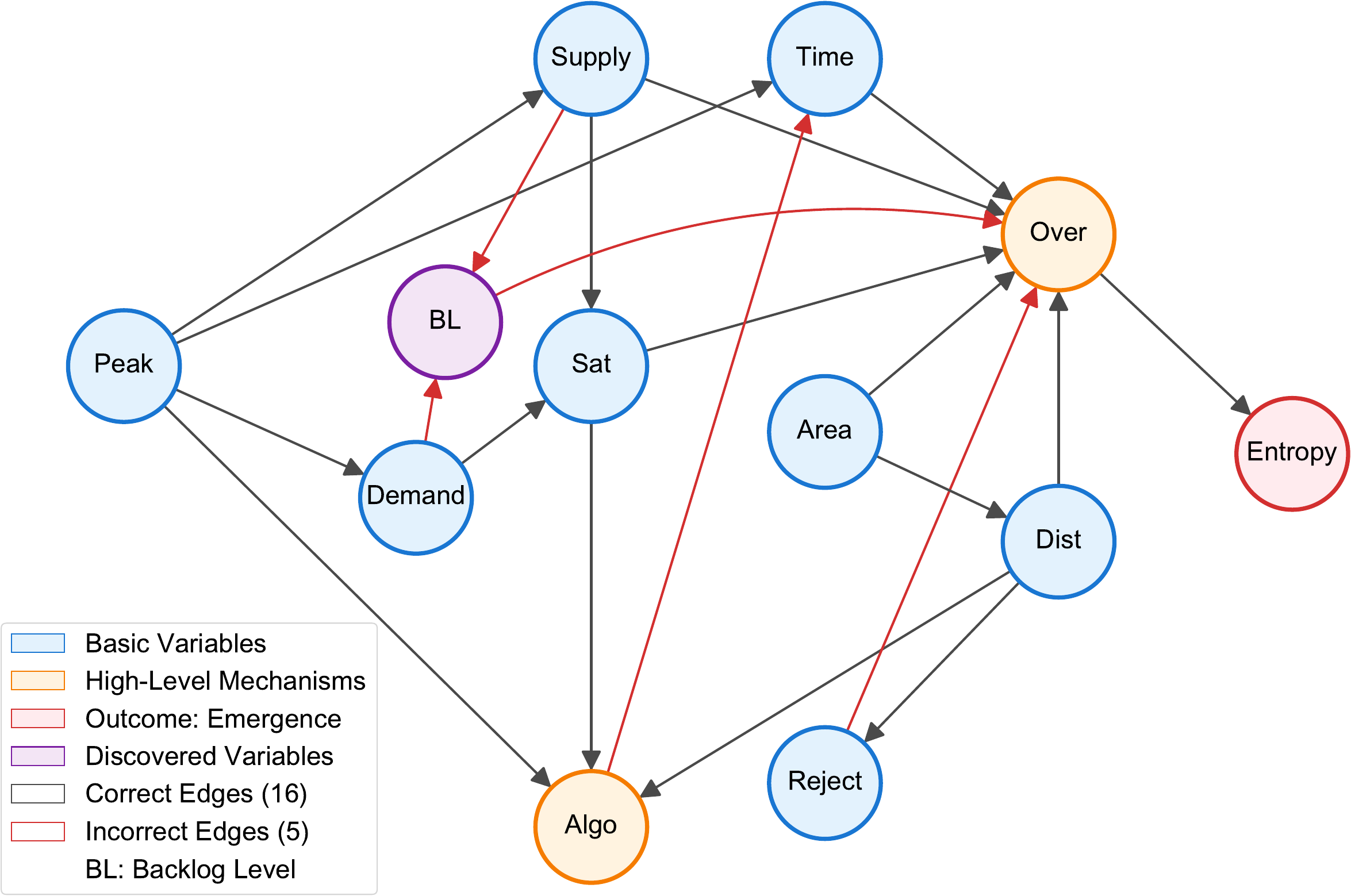}
  \caption{w/o A3-add}
  \label{fig:ablation-exp2}
\end{subfigure}

\vspace{0.6em}

\begin{subfigure}[t]{0.48\textwidth}
  \centering
  \includegraphics[width=\linewidth]{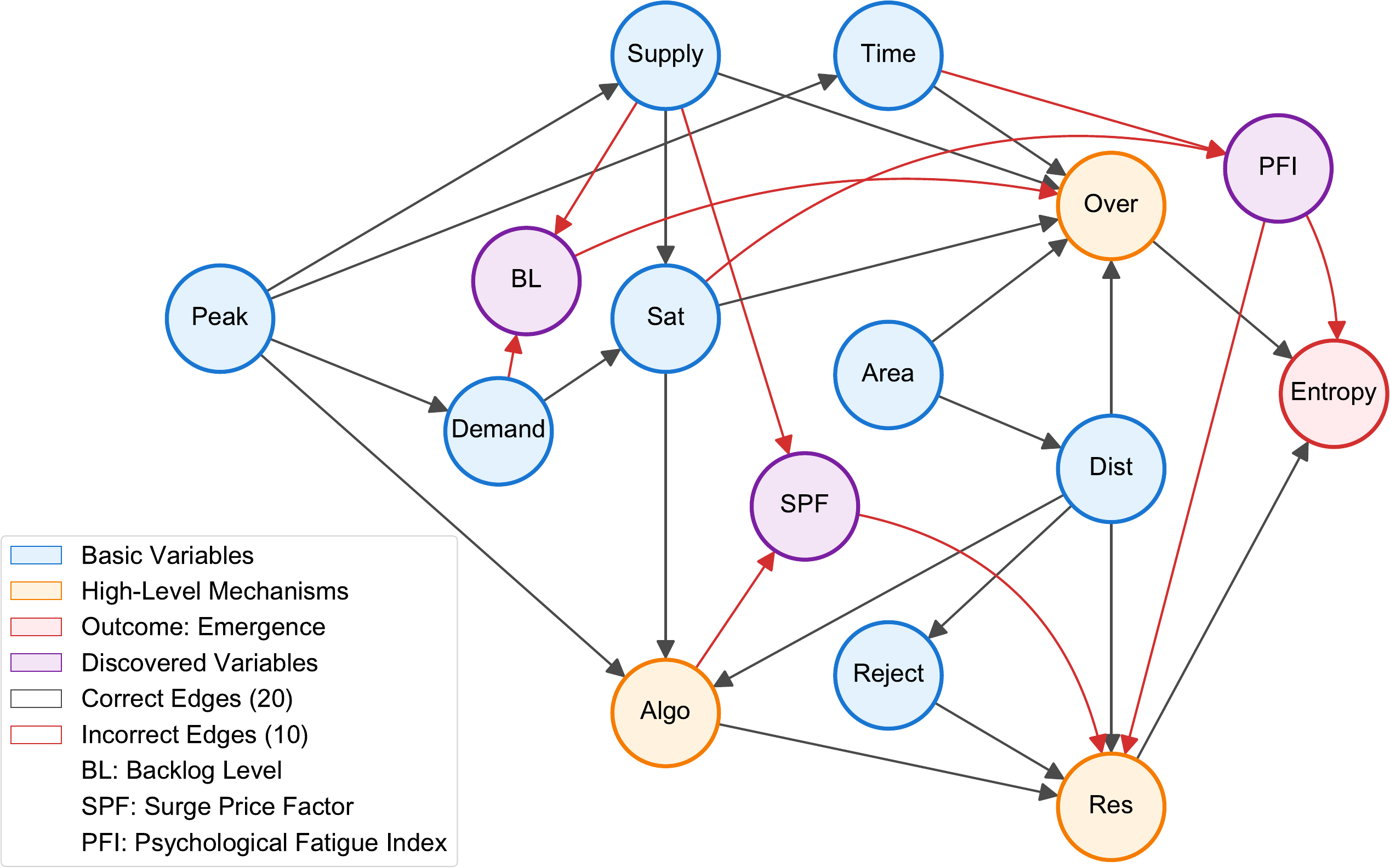}
  \caption{w/o A5 (no counterfactual)}
  \label{fig:ablation-exp3}
\end{subfigure}\hfill
\begin{subfigure}[t]{0.48\textwidth}
  \centering
  \includegraphics[width=\linewidth]{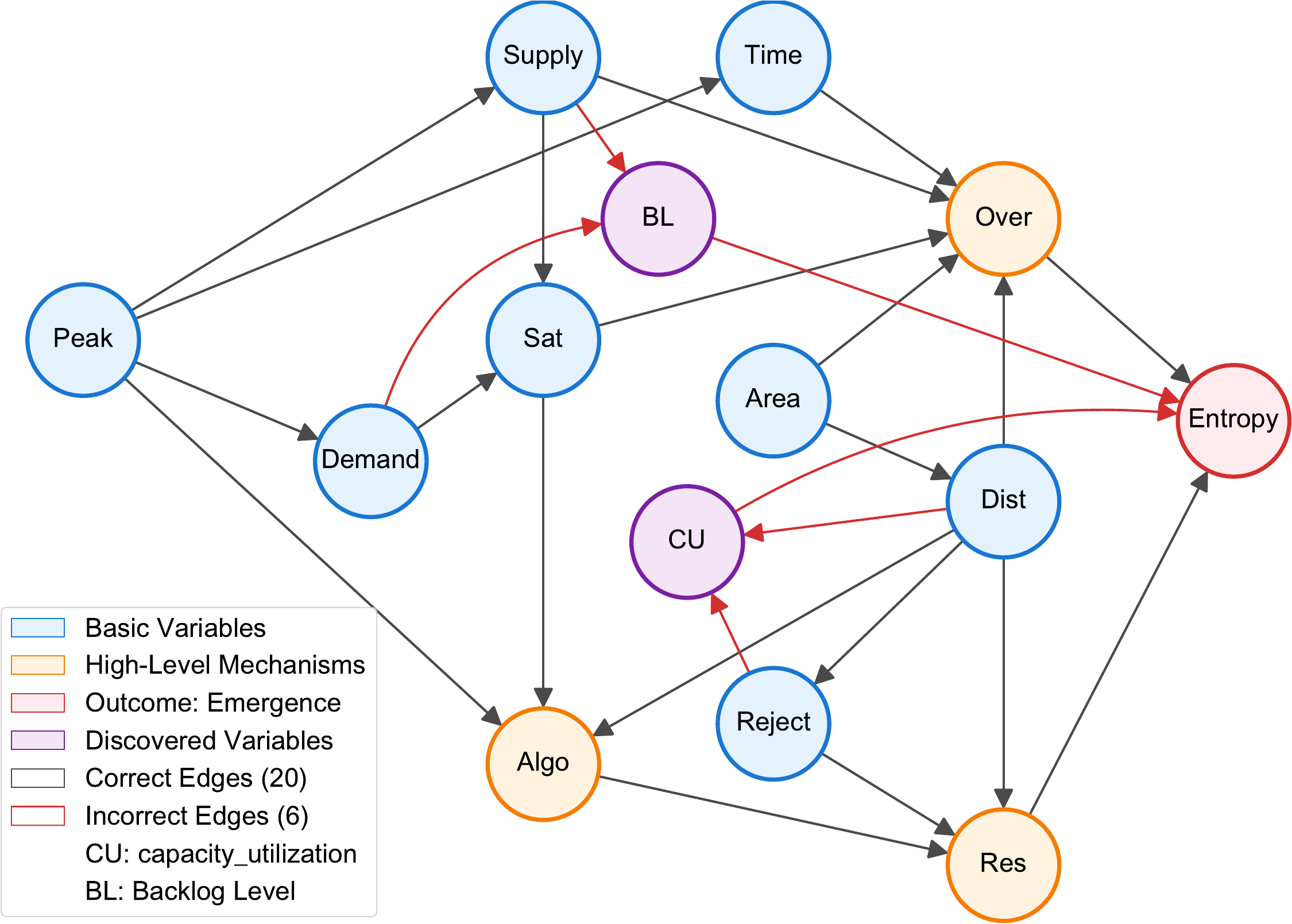}
  \caption{w/o competing hypotheses }
  \label{fig:ablation-exp4}
\end{subfigure}

\caption{Ablation study on the O2O delivery simulation.}
\label{fig:ablation-o2o}
\end{figure*}

\subsection{Analysis of Estimated Refinement Effectiveness $(p,C)$}
\label{app:pc_analysis}

Figure~\ref{fig:pc_scatter} summarizes the estimated refinement-effectiveness parameters $(p,C)$ across LLM backbones.
Recall that $p$ captures the probability that an iteration yields an informative refinement step, while $C$ measures the typical
(relative) contraction strength conditioned on being informative; together, they govern the expected geometric decay rate through
the product $pC$ (Theorem~\ref{thm:geom}).

We observe clear backbone-dependent differences that are consistent with the refinement dynamics in
Figure~\ref{fig:add_prune_dynamics}. DeepSeek-V3.2 exhibits the strongest overall effectiveness, with both a relatively large $p$
and a strong conditional contraction, resulting in the largest estimated $pC$ and thus the fastest expected decay of residual
dependence. GPT-5 mini follows with a similarly balanced profile, suggesting that informative steps occur frequently and yield
reliable contraction.

Qwen3-235B-A22B shows a comparatively larger $C$ but a smaller $p$, indicating more intermittent progress: when refinement succeeds
it tends to contract strongly, but successful steps occur less often. This is consistent with the add/prune curves where the
candidate set is pruned effectively yet improvements are not uniformly sustained across rounds. DeepSeek-R1-Distill-Qwen-32B
exhibits a smaller $p$ (with moderate $C$), implying that informative refinement steps are rarer, aligning with its slower
stabilization behavior over rounds.

Finally, Gemma3-27B yields a smaller $pC$, indicating less consistent contraction overall. Notably, this does not preclude early
stabilization of the estimated Markov-boundary size in the add/prune curves; rather, it suggests that subsequent refinement is less
reliably contractive (e.g., weaker regularization of the broader candidate set), which can delay stabilization in terms of residual
dependence.

The concrete estimation procedure for $(p,C)$ from the empirical $\{F_t\}$ trajectories is provided in
Appendix~\ref{app:pc_estimation}.

\subsection{Ablation Design and Analysis}
\label{app:ablation}

\paragraph{Ablation design.}
We ablate four core components of \textsc{CAMO} while keeping the rest of the pipeline unchanged.
\textbf{w/o A1\&A2} removes the worldview parsing and integration stages, so the system runs without
structured prior hypotheses about variables and relations.
\textbf{w/o A3-add} disables the add step in the A2--A3 refinement loop, i.e., factors are only
pruned/refined without explicitly re-introducing missed candidates.
\textbf{w/o A5 (no counterfactual)} removes the counterfactual adjudication module, so candidate
edges/hypotheses are not filtered by intervention-based checks.
\textbf{w/o competing hypotheses} disables hypothesis competition, replacing set-based competing
hypotheses with a single-pass selection to test whether explicit competition is necessary.
All variants are evaluated under the same simulator setting and budget as the full model.

\paragraph{Analysis.}Table~\ref{tab:ablation} shows that each module supports either \emph{coverage} of the true
upstream ancestry or \emph{compactness} of the learned structure. Removing A1\&A2 causes factor
discovery to fail, indicating that worldview parsing and integration are necessary to surface
causally relevant factors. Disabling A3-add weakens both local (MB) and upstream recovery, suggesting
the add step is important for bringing back missed-but-necessary causes. Removing A5 (no
counterfactual) largely preserves the local interface but introduces more spurious structure,
showing that counterfactual adjudication primarily acts as a redundancy filter.

\section{Implementation Details}
\label{app:imp}

\subsection{Estimating Refinement-Capability Parameters}
\label{app:pc_estimation}

This appendix describes how we compute practical diagnostics $(\hat p,\hat C)$
from the A3 refinement trajectory. These quantities are meant to be simple,
interpretable summaries of refinement effectiveness, not statistically exact
estimators of the theoretical $(p,C)$ in Theorem~\ref{thm:geom}.

\subsubsection{A computable uncertainty score $H_b$}
\label{app:hb}

For any variable set $U$, we use $H_b(Y\mid U)$ to denote a \emph{computable}
score of how uncertain $Y$ remains after conditioning on $U$. In practice,
$H_b(Y\mid U)$ is estimated by a held-out predictive loss of a classifier/regressor
that predicts $Y$ from $U$ (e.g., cross-entropy for discrete $Y$). Smaller
$H_b(Y\mid U)$ means $U$ explains/predicts $Y$ better.

\subsubsection{Information gain and factor admission}
\label{app:information_gain}

At refinement step $t$, let $V^{(t)}$ be the current admitted factor set. For a
candidate factor $Z$, we define its \emph{computable information gain} as
\begin{equation}
\label{eq:app_deltaIc_def_rewrite}
\Delta^{c} I(Z)
\;:=\;
H_b(Y \mid V^{(t)}) - H_b(Y \mid V^{(t)}, Z).
\end{equation}
This measures how much adding $Z$ reduces the residual uncertainty of $Y$ beyond
what $V^{(t)}$ already explains. We admit $Z$ if
\begin{equation}
\label{eq:app_admission_rule_rewrite}
\Delta^{c} I(Z) > \tau_I,
\end{equation}
where $\tau_I\ge 0$ is a small margin (default $\tau_I=0$).

\paragraph{Targeted evaluation.}
When estimating $H_b$ and $\Delta^c I(\cdot)$, we follow~\citet{liu2024discovery}
and evaluate them on the subset of records that is hardest to predict under the
current $V^{(t)}$ (the ``worst-explained'' group). This prevents improvements on
easy cases from masking failures on hard cases.

\subsubsection{Proxy residual difficulty $\widehat F_t$}

Theorem~\ref{thm:geom} analyzes a residual dependence quantity $F_t$ that is not
directly computable. Along the refinement trajectory, we track a computable
surrogate,
\begin{equation}
\label{eq:app_Fhat_def_rewrite}
\widehat F_t
\;:=\;
H_b(Y \mid V^{(t)}),
\end{equation}
where $H_b$ is evaluated using the same protocol as above (i.e., on the
worst-explained group at step $t$). Smaller $\widehat F_t$ indicates that the
current admitted set explains $Y$ better in the hardest region.

\subsubsection{Successful steps and estimating $\hat p$ and $\hat C$}

Because $\widehat F_t$ is measured with finite samples, we introduce a small
tolerance $\epsilon>0$ and call step $t$ \emph{successful} if the proxy difficulty
decreases by at least $\epsilon$:
\begin{equation}
\label{eq:app_success_simple_rewrite}
\widehat F_{t+1} \le \widehat F_t - \epsilon.
\end{equation}

\paragraph{Success frequency $\hat p$.}
We estimate $\hat p$ as the fraction of successful refinement steps:
\begin{equation}
\label{eq:app_p_hat_rewrite}
\hat p
=
\frac{1}{T}
\sum_{t=0}^{T-1}
\mathbf 1\!\left\{
\widehat F_{t+1} \le \widehat F_t - \epsilon
\right\}.
\end{equation}

\paragraph{Typical contraction $\hat C$.}
On successful steps, we measure the relative decrease
\begin{equation}
\label{eq:app_rt_rewrite}
r_t
=
\frac{\widehat F_t - \widehat F_{t+1}}{\widehat F_t},
\end{equation}
and estimate the typical contraction as
\begin{equation}
\label{eq:app_C_hat_rewrite}
\hat C
=
\operatorname{median}
\{\, r_t \mid \widehat F_{t+1} \le \widehat F_t - \epsilon \,\}.
\end{equation}

\subsubsection{Interpretation}

$\hat p$ answers \emph{how often} refinement produces a meaningful reduction in
proxy residual difficulty, while $\hat C$ summarizes \emph{how large} that
reduction typically is when it happens. Their product $\hat p\hat C$ provides a
simple summary of empirical refinement effectiveness along the A3 trajectory.

\subsection{Decision Worldview Scoring and Selection}
\label{app:judge}

To score and select candidate decision worldviews at iteration $k$, A2
uses a lightweight two-stage LLM-judge procedure adapted from \citet{zheng2023judging}.

\paragraph{Stage I: rubric scoring.}
Given $n$ candidates $\{W_i\}$, the judge assigns each $W_i$ a 1--10 score under
a fixed rubric (e.g., feasibility, novelty, rigor, and target alignment; with
optional weights). A2 keeps only the top $m$ candidates (typically $m\in\{3,4\}$)
to avoid exhaustive $\binom{n}{2}$ comparisons.

\paragraph{Stage II: pairwise comparison.}
On the shortlist, the judge performs pairwise comparisons and outputs a winner
(or tie) with a brief justification. A2 aggregates outcomes into a ranking and
selects the working set; if preference cycles occur, A2 runs one additional
multi-way adjudication on the tied set to obtain a stable order. The final
decision worldview $W^{(k)}$ is chosen from the working set.

\subsection{Edge Prioritization for A4}
\label{app:a4_ucb}

This appendix specifies a simple, fully computable heuristic used by A4
to prioritize unresolved edges for simulator-based interventions.

\paragraph{Inputs from A3.}
Let $\mathcal{E}_u$ denote the set of unresolved edges after observational
discovery (e.g., ambiguous orientations in a CPDAG/PAG). For each $e\in\mathcal{E}_u$,
A3 provides an importance score $w_e\in[0,1]$ and an uncertainty score
$u_e\in[0,1]$. Intuitively, $w_e$ measures how consequential resolving $e$ is for
downstream causal accuracy, while $u_e$ measures how weakly $e$ is identified
from observational constraints.

\paragraph{Importance and uncertainty signals.}
The importance score $w_e$ is derived from A3's stabilized representation, for
example using (i) association with the target outcome, (ii) frequency of $e$
across repeated discovery runs. The
uncertainty score $u_e$ is derived from the learned CPDAG/PAG; in a minimal
implementation, $u_e=1$ if $e$ is not fully oriented (or contains ambiguous PAG
marks) and $u_e=0$ otherwise.

Optionally, A3 may provide a stability signal $\hat{\sigma}_e\in[0,1]$ by
re-running observational discovery on subsampled observations and quantifying
orientation variability. One simple choice is
$\hat{\sigma}_e = 1 - \mathrm{agree}(e)$, where $\mathrm{agree}(e)$ is the
fraction of subsamples that return the same orientation for $e$. 

\paragraph{Optimism-weighted scoring.}
Given $(w_e,u_e)$ and, when available, $\hat{\sigma}_e$, A4 ranks edges by
the UCB-inspired heuristic
\begin{equation}
\label{eq:app_a4_ucb}
\mathrm{score}(e) \;=\; w_e\big(u_e + \beta\,\hat{\sigma}_e\big),
\end{equation}
where $\beta>0$ controls the weight on stability-based exploration. If no
stability estimate is available, we set $\hat{\sigma}_e=0$, yielding
$\mathrm{score}(e)=w_e u_e$.

A4 selects the top-$K$ edges by $\mathrm{score}(e)$ and generates
intervention scripts accordingly.

\section{Experimental Details and Metrics}
\label{app:metrics}




\subsection{Factor Discovery Metrics}
\label{app:metrics:factor}

We define factor discovery metrics based on the ground-truth causal graph of the
simulator. Let $Y$ denote the target variable, $B(Y)$ its Markov boundary, and
$\mathrm{An}(Y)$ the set of all causal ancestors of $Y$ (excluding $Y$ itself).
Given a set of proposed high-level factors $S$, we categorize factors as:
\begin{itemize}
  \item \textbf{MB}: $S \cap B(Y)$, factors belonging to the Markov boundary;
  \item \textbf{AN}: $S \cap (\mathrm{An}(Y)\setminus B(Y))$, factors that are
  causal ancestors but not in the boundary;
  \item \textbf{OT}: $S \setminus \mathrm{An}(Y)$, off-target or spurious factors.
\end{itemize}

\paragraph{MB recovery.}
We report the counts of MB, AN, and OT factors. In addition, we compute Precision,
Recall, and F1 with respect to the Markov boundary $B(Y)$:
\begin{equation}
\label{eq:mb-metrics}
\begin{aligned}
\mathrm{Precision}_{\mathrm{MB}} &= \frac{|S \cap B(Y)|}{|S|}, \\
\mathrm{Recall}_{\mathrm{MB}} &= \frac{|S \cap B(Y)|}{|B(Y)|}, \\
\mathrm{F1}_{\mathrm{MB}} &= \frac{2\,\mathrm{Precision}_{\mathrm{MB}}\,\mathrm{Recall}_{\mathrm{MB}}}
{\mathrm{Precision}_{\mathrm{MB}} + \mathrm{Recall}_{\mathrm{MB}}}.
\end{aligned}
\end{equation}

\paragraph{Ancestor F1.}
To assess recovery of ancestral relations with respect to the target variable
$Y$, we evaluate \emph{reachability} (transitive) relations rather than ancestor
node-set overlap. Let $\mathrm{TC}(G)$ denote the transitive closure of $G$.
We define the target-specific ancestor-relation set
\begin{equation}
\label{eq:ancY-def}
\begin{aligned}
R_Y(G)
&=\{(u,Y)\in V^{\mathrm{obs}}\times V^{\mathrm{obs}}:\; u\neq Y, \\
&\qquad (u,Y)\in \mathrm{TC}(G)\}.
\end{aligned}
\end{equation}
Let $R_Y^\ast = R_Y(G^\ast)$ and $\widehat{R}_Y = R_Y(\hat{G})$. We compute
precision/recall/F1 over these ordered pairs:
\begin{equation}
\begin{aligned}
\mathrm{Precision}_{\mathrm{anc}} &=
\frac{|\widehat{R}_Y\cap R_Y^\ast|}{|\widehat{R}_Y|}, \\
\mathrm{Recall}_{\mathrm{anc}} &=
\frac{|\widehat{R}_Y\cap R_Y^\ast|}{|R_Y^\ast|}, \\
\mathrm{Anc\text{-}F1} &=
\frac{2\,\mathrm{Precision}_{\mathrm{anc}}\,\mathrm{Recall}_{\mathrm{anc}}}
{\mathrm{Precision}_{\mathrm{anc}}+\mathrm{Recall}_{\mathrm{anc}}}.
\end{aligned}
\end{equation}

This metric credits a method if it recovers the correct \emph{transitive}
upstream influence on $Y$ (e.g., $A\!\to\!B\!\to\!Y$ implies $(A,Y)\in \mathrm{TC}$),
regardless of whether the influence is realized by a direct edge or a multi-hop
path in the predicted graph.

\subsection{Projection to Observed-Variable Space}
\label{app:metrics:projection}

CAMO may introduce computable factor nodes (constructed variables) that
do not exist in the raw log space. To fairly compare against baselines that
operate only on observed variables, we evaluate graph-structure metrics on the
observed-variable space by projecting any learned graph $G$ onto
$\mathcal{V}^{\mathrm{obs}}$.

\paragraph{Observed variables and factor supports.}
Let $\mathcal{V}^{\mathrm{obs}}$ be the set of logged (raw) variables.
Each constructed factor node $Z$ is associated with a \emph{support set}
$\mathrm{supp}(Z)\subseteq \mathcal{V}^{\mathrm{obs}}$, defined as the set of
raw variables used by the deterministic construction rule of $Z$
(e.g., counts, ratios, summary statistics, graph metrics). For raw variables
$X\in\mathcal{V}^{\mathrm{obs}}$, we define $\mathrm{supp}(X)=\{X\}$.

\paragraph{Projection operator.}
Given a learned graph $G=(V,E)$ that may include constructed factors, we define
the projected graph $\Pi(G)=(\mathcal{V}^{\mathrm{obs}},E_{\Pi})$ by removing all
non-observed nodes and \emph{lifting} each edge to observed supports. For every
directed or partially oriented edge $(U \rightarrow V)\in E$, we add edges from
$\mathrm{supp}(U)$ to $\mathrm{supp}(V)$:
\begin{equation}
\begin{aligned}
E_{\Pi}
&:= \bigcup_{(U\rightarrow V)\in E}
\Bigl\{\, x \rightarrow z \;:\;
x\in\mathrm{supp}(U), \\
&\hspace{3.8em}
z\in\mathrm{supp}(V)
\,\Bigr\}.
\end{aligned}
\end{equation}
Note that when both $U$ and $V$ are observed variables, the projection reduces to
the original edge between them. If the learned structure is a CPDAG/PAG with an
ambiguous endpoint on $(U\,\circ\!\!-\!\!\circ\,V)$, we project it similarly while
preserving ambiguity:
\begin{equation}
\begin{aligned}
\Pi\bigl(U\,\circ\!\!-\!\!\circ\,V\bigr)
&=
\Bigl\{\, x\,\circ\!\!-\!\!\circ\,z \;:\;
x\in\mathrm{supp}(U), \\
&\hspace{3.8em}
z\in\mathrm{supp}(V)
\,\Bigr\}.
\end{aligned}
\end{equation}
Finally, we simplify $\Pi(G)$ by removing duplicate edges. Self-loops
($x\rightarrow x$) are discarded.

\paragraph{Rationale.}
This projection attributes dependencies mediated by constructed factors back to
the raw variables from which they are computed, enabling direct comparison with
baselines that cannot introduce factor nodes. All graph-level structure metrics
(e.g., SHD, NHD, Added/Missed/Reversed edges) are computed on $\Pi(G)$.

\begin{table*}[!t]
\centering
\small
\setlength{\tabcolsep}{5pt}
\renewcommand{\arraystretch}{1.12}
\caption{AgentSociety controllable mechanisms as causal nodes (one row per node).}
\label{tab:rq2_agentsociety_nodes}
\begin{tabular}{p{0.34\linewidth} p{0.58\linewidth}}
\toprule
Node $v$ & Controllable mechanism (simulator knob) \\
\midrule
$\mathrm{ExposurePolicyType}$ & Recommendation/exposure policy family. \\
$\mathrm{ExposureBiasLevel}$ & Strength of homophilous exposure bias. \\
$\mathrm{NetworkTopologyType}$ & Clustered vs.\ mixing interaction topology. \\
$\mathrm{PersuasiveMsgSupplyIntensity}$ & Supply intensity of persuasive messages. \\
$\mathrm{InitialOpinionShift}$ & Shift of initial opinion distribution. \\
$\mathrm{InflammatoryPostingRate}$ & Rate of inflammatory content injection. \\
$\mathrm{ModerationPolicyType}$ & Moderation on/off policy family. \\
$\mathrm{ModerationAggressiveness}$ & Moderation strictness / action threshold. \\
\bottomrule
\end{tabular}
\end{table*}

\begin{table*}[!t]
\centering
\small
\setlength{\tabcolsep}{5pt}
\renewcommand{\arraystretch}{1.12}
\caption{Smallville controllable mechanisms as causal nodes (one row per node).}
\label{tab:rq2_smallville_nodes}
\begin{tabular}{p{0.34\linewidth} p{0.58\linewidth}}
\toprule
Node $v$ & Controllable mechanism (direct input edit) \\
\midrule
$\mathrm{SeedInfoCoverage}$ & fraction of agents receiving the coordination seed. \\
$\mathrm{SeedInfoSalience}$ & salience/strength of injected seed memory. \\
$\mathrm{PublicAnnouncement}$ & presence of a shared public bulletin/notice. \\
$\mathrm{EncounterBudget}$ & encounter/opportunity budget for interactions. \\
$\mathrm{KeyAgentSeeding}$ & whether hub/central agents are seeded first. \\
\bottomrule
\end{tabular}
\end{table*}

\subsection{Causal Structure Recovery Metrics}
\label{app:metrics:structure}

We evaluate causal structure recovery by comparing the projected learned graph
$\hat{G}=\Pi(G)$ with the ground-truth graph $G^{\ast}$ in the observed-variable
space (Appendix~\ref{app:metrics:projection}). Let
$G^{\ast}=(V^{\mathrm{obs}},E^{\ast})$ and
$\hat{G}=(V^{\mathrm{obs}},\hat{E})$.

\paragraph{Directed edge metrics.}
Following prior work, we evaluate edge-wise performance on the set of
\emph{directed edges}. Let $E^{\ast}\subset V^{\mathrm{obs}}\times V^{\mathrm{obs}}$
and $\hat{E}\subset V^{\mathrm{obs}}\times V^{\mathrm{obs}}$ denote the sets of
ground-truth and predicted directed edges, respectively. We define
\begin{equation}
\begin{aligned}
\mathrm{TP} &= |\hat{E}\cap E^{\ast}|, \\
\mathrm{FP} &= |\hat{E}\setminus E^{\ast}|, \\
\mathrm{FN} &= |E^{\ast}\setminus \hat{E}|, \\
\mathrm{TN} &= |V^{\mathrm{obs}}|(|V^{\mathrm{obs}}|-1)
              -\mathrm{TP}-\mathrm{FP}-\mathrm{FN}.
\end{aligned}
\end{equation}
Based on these counts, we compute
\begin{equation}
\begin{aligned}
\mathrm{Precision} &= \frac{\mathrm{TP}}{\mathrm{TP}+\mathrm{FP}}, \\
\mathrm{Recall} &= \frac{\mathrm{TP}}{\mathrm{TP}+\mathrm{FN}}, \\
\mathrm{F1} &=
\frac{2\,\mathrm{Precision}\,\mathrm{Recall}}
{\mathrm{Precision}+\mathrm{Recall}}.
\end{aligned}
\end{equation}

\paragraph{Accuracy.}
We additionally report the accuracy (Acc) over directed edges, following the
edge-set evaluation in our implementation. Let $n = |V^{\mathrm{obs}}|$ and
consider all ordered pairs $(u,v)$ with $u\neq v$ as candidate directed edges.
Using $\mathrm{TP},\mathrm{FP},\mathrm{FN}$ as defined above, we set
\begin{equation}
\mathrm{TN} = n(n-1) - \mathrm{TP} - \mathrm{FP} - \mathrm{FN},
\end{equation}
and compute
\begin{equation}
\mathrm{Acc} = \frac{\mathrm{TP}+\mathrm{TN}}{\mathrm{TP}+\mathrm{TN}+\mathrm{FP}+\mathrm{FN}}
= \frac{\mathrm{TP}+\mathrm{TN}}{n(n-1)}.
\end{equation}

\paragraph{False positive rate.}
We additionally report the false positive rate (FPR) over directed edges:
\begin{equation}
\mathrm{FPR} = \frac{\mathrm{FP}}{\mathrm{FP}+\mathrm{TN}}.
\end{equation}

\paragraph{SHD and error decomposition.}
Let $S(G)$ denote the (undirected) skeleton of a graph $G$, i.e.,
$S(G)=\{\{u,v\}:(u\!\to\!v)\in E(G)\ \text{or}\ (v\!\to\!u)\in E(G)\}$.
We decompose the structural Hamming distance (SHD) into four components:
\begin{itemize}
  \item \textbf{Added}: $|S(\hat{G}) \setminus S(G^{\ast})|$, extra adjacencies;
  \item \textbf{Missed}: $|S(G^{\ast}) \setminus S(\hat{G})|$, missing adjacencies;
  \item \textbf{Reversed}: common adjacencies whose orientation in $\hat{G}$ is the
  opposite of $G^{\ast}$;
  \item \textbf{Unoriented}: common ground-truth adjacencies whose direction is not
  resolved in $\hat{G}$ (e.g., \texttt{o--o} or \texttt{---} in CPDAG/PAG outputs).
\end{itemize}
We define
\begin{equation}
\mathrm{SHD}=\mathrm{Added}+\mathrm{Missed}+\mathrm{Reversed}+\mathrm{Unoriented}.
\end{equation}

\paragraph{Ancestor metrics.}
The computation of Anc-F1 is described in Appendix~\ref{app:metrics:factor}.

\section{Interventional Ranking without Ground-Truth Graphs}
\label{app:metrics:intervention}

This appendix specifies the evaluation protocol when no ground-truth causal
graph is available. The goal is to test whether a learned graph provides
\emph{actionable guidance} by ranking effective simulator-supported interventions
ahead of ineffective (placebo) ones.

\subsection{Pre-registered Intervention Pool and Success Labels}
\label{app:metrics:intervention_pool}

For each environment--target pair, we pre-register a fixed pool of
simulator-supported intervention scripts
\begin{equation}
S = \{s_1,\dots,s_{|S|}\}, \quad |S|=12,
\end{equation}
where each script modifies exactly one controllable mechanism while keeping
other settings unchanged. To increase discriminability, $S$ intentionally
contains many \emph{placebo/weak-effect} scripts (small-magnitude changes,
counteracting changes, or mechanism-irrelevant toggles), so that the set of
effective scripts is sparse.

Each script is executed under multiple random seeds. A script is labeled
\emph{successful} if it induces a statistically significant change in the
emergent outcome $Y$ relative to a no-intervention baseline:
\begin{equation}
\mathrm{success}(s)\in\{0,1\}.
\end{equation}
We denote the set of successful scripts by
\begin{equation}
S^+ = \{s\in S:\mathrm{success}(s)=1\}.
\end{equation}
Crucially, $S^+$ is obtained from rollouts and is \emph{method-independent}:
no method is given $S^+$; all methods only output a learned graph.

\subsection{From a Causal Graph to an Intervention Ranking (RWR)}
\label{app:metrics:rwr}

Methods output a directed causal graph $G=(V,E)$ but do not directly output a
ranking over scripts. We derive a comparable ranking using Random Walk with
Restart (RWR), propagating influence \emph{from the target back to upstream
mechanisms}.

\paragraph{Script-to-node mapping.}
Each script $s$ targets a mechanism node $v(s)$ defined in the intervention
tables below. If $v(s)\notin V$ for a learned graph $G$, we assign score $0$ to
$s$ (ties are broken deterministically).

\paragraph{Upstream propagation.}
Let $A$ be the row-normalized adjacency matrix of $G$ restricted to nodes
connected to $Y$. We run RWR on the reversed flow:
\begin{equation}
\label{eq:rwr}
\pi = \alpha e_Y + (1-\alpha) A^\top \pi,
\end{equation}
where $e_Y$ is one-hot on $Y$ and $\alpha\in(0,1)$ is the restart probability.
We rank scripts by $\pi(v(s))$ in descending order.

\subsection{Ranking Metrics}
\label{app:metrics:ranking_metrics}

We report ranking quality at $K=5$ (a short actionable shortlist). Let $s_{(i)}$
be the $i$-th ranked script.

\paragraph{Precision@K.}
\begin{equation}
\mathrm{P@K}=\frac{1}{K}\sum_{i=1}^{K}\mathrm{success}(s_{(i)}).
\end{equation}

\paragraph{MAP@K.}
\begin{equation}
\label{eq:mapk}
\begin{aligned}
\mathrm{MAP@K}
&=
\frac{1}{\min(K,|S^+|)}
\sum_{i=1}^{K} \\
&\Bigl(\mathrm{P@i}\Bigr)\cdot
\mathbf{1}\!\left[\mathrm{success}(s_{(i)})=1\right].
\end{aligned}
\end{equation}

\paragraph{MRR.}
\begin{equation}
\mathrm{MRR}
=
\frac{1}{\min\{i:\mathrm{success}(s_{(i)})=1\}}.
\end{equation}

\subsubsection{AgentSociety: Polarization and Inflammatory-Message Spread}
\label{app:rq2_agentsociety}

We study two emergent targets in AgentSociety: \emph{opinion polarization} and
the \emph{spread of inflammatory messages}. 
For reference, the manually annotated root causes are provided in Table~\ref{tab:rq2_agentsociety_nodes}.

\subsubsection{Smallville: Agent Coordination}
\label{app:rq2_smallville}

We evaluate on Smallville with the emergent target \emph{agent coordination}.
We use only simulator-realizable interventions implemented as direct edits to
memories/public information/encounter opportunities (no hand-crafted behavior
rules). For reference, the manually annotated root causes are provided in Table~\ref{tab:rq2_smallville_nodes}.

\section{Baselines}
\label{app:baselines_prompts}

\subsection{Factor Discovery Baselines}
\label{app:factor_baselines}

This appendix lists prompt templates for factor-discovery baselines used in
\textsection~\ref{sec:experiments}. All baselines use the same simulator
description and target outcome $Y$, and output a set of candidate factors $S$
in the same format as CAMO. Unless stated otherwise, baselines use no external
causal feedback (e.g., CI tests, graph feedback, or interventions) and no post
hoc pruning beyond the model's own output.

\paragraph{LLM-Text (single-round).}
Given only the simulator description and task context, the LLM proposes a list
of high-level candidate factors for $Y$ in one shot. The returned list is used
directly as $S$.

\paragraph{LLM-Data (single-round).}
Same as LLM-Text, but additionally provides a small observational data. The one-shot output list is
used directly as $S$.

\paragraph{LLM-CoT (single-round, text-only).}
Same inputs as LLM-Text (no observational data), but the LLM is instructed to
reason before listing factors. The final list is used as $S$.

\paragraph{LLM-MultiRound (multi-round, text-only).}
Same inputs as LLM-Text (no observational data). The LLM iteratively revises the
factor list across multiple rounds using only the dialogue history. No external
verification signal is provided, and the final round's list is taken as $S$.

\subsection{Factor Discovery Baseline Prompts}
\label{app:factor_prompts}

\subsubsection{Text-only Prompt}

\begin{tcolorbox}[colback=gray!10,colframe=gray!60,boxrule=0.6pt,arc=2pt]
\textbf{Prompt.}
You are given: (1) a description of a multi-agent simulation environment and
(2) an outcome of interest $Y$.

Propose a compact set of high-level factors that could causally influence $Y$.
Each factor should be a nonlinear, computable abstraction over collective agent
behavior (e.g., aggregation effects, feedback loops, distributional patterns),
not a single low-level action or implementation detail.

Output a list of factors. For each factor, provide a short name and one sentence
describing how it could affect $Y$.
\end{tcolorbox}

\subsubsection{Data-grounded Prompt}

\begin{tcolorbox}[colback=gray!10,colframe=gray!60,boxrule=0.6pt,arc=2pt]
\textbf{Prompt.}
You are given: (1) a description of a multi-agent simulation environment,
(2) observational summaries from simulation runs (e.g., samples and/or summary
statistics), and (3) an outcome of interest $Y$.

Using the observed data, propose a compact set of high-level, computable factors
that may causally influence $Y$. Each factor should be supported by or consistent
with the observed patterns.

Output a list of factors. For each factor, provide a short name and one sentence
describing how it could affect $Y$.
\end{tcolorbox}

\subsubsection{Chain-of-Thought Prompt (text-only)}

\begin{tcolorbox}[colback=gray!10,colframe=gray!60,boxrule=0.6pt,arc=2pt]
\textbf{Prompt.}
You are given: (1) a description of a multi-agent simulation environment and
(2) an outcome of interest $Y$.

First, briefly reason step by step about which collective properties of agent
behavior could plausibly influence $Y$. Then propose a compact set of high-level,
computable factors.

Output only the final factor list. For each factor, provide a short name and one
sentence describing how it could affect $Y$.
\end{tcolorbox}

\subsubsection{Multi-round Prompt (text-only, no causal feedback)}

\begin{tcolorbox}[colback=gray!10,colframe=gray!60,boxrule=0.6pt,arc=2pt]
\textbf{Round 1 Prompt.}
You are given: (1) a description of a multi-agent simulation environment and
(2) an outcome of interest $Y$.

Propose an initial compact set of high-level, nonlinear, computable factors that
may causally influence $Y$. For each factor, provide a short name and one
sentence description.
\end{tcolorbox}

\begin{tcolorbox}[colback=gray!10,colframe=gray!60,boxrule=0.6pt,arc=2pt]
\textbf{Round $t>1$ Prompt.}
Here is your current factor list. Revise it using only this conversation:
merge near-duplicates, clarify ambiguous items, and optionally add missing
high-level factors. Keep the final list compact and interpretable.

Output only the updated factor list (name + one sentence per factor).
\end{tcolorbox}

\subsection{Causal Structure Recovery Baselines}
\label{app:structure_base}

We compare against three classes of methods.

\emph{Statistical causal discovery (SCD)} methods include
PC~\cite{spirtes1991algorithm}, FCI~\cite{spirtes2000causation},
GES~\cite{chickering2002optimal} and MMHC~\cite{tsamardinos2006max},
all applied directly to observed variables.
PC is a constraint-based procedure that uses conditional-independence tests to
learn an equivalence class of DAGs (typically returned as a CPDAG) under causal
sufficiency; FCI extends PC to allow latent confounders and selection bias,
returning a PAG with possibly undirected/partially oriented edges.
GES is a score-based greedy search over Markov equivalence classes that adds and
then deletes edges to optimize a decomposable score (e.g., BIC), producing a
CPDAG; MMHC is a hybrid approach that first performs constraint-based neighbor
selection (max--min parents-and-children) to prune candidates, followed by a
score-based hill-climbing phase to orient/refine edges.

\emph{Pure LLM methods} include Efficient-CDLMs~\cite{jiralerspong2024efficient},
MAC~\cite{le2024multi}, and PAIRWISE~\cite{kiciman2023causal}.
Efficient-CDLMs employs a BFS-style prompting strategy that leverages the DAG
structure to construct causal graphs with fewer LLM queries than exhaustive
pairwise prompting.
MAC formulates causal discovery as a multi-agent debate among LLMs, where agents
argue for/against candidate relations and the final graph is obtained by
aggregating debated judgments.
PAIRWISE queries an LLM for pairwise causal direction decisions and aggregates
the resulting judgments into a global graph without relying on explicit
statistical tests, potentially leaving relations uncertain when evidence is weak.

Finally, \emph{hybrid SCD+LLM methods} refine an initial SCD-produced graph using
LLM reasoning, including SCD-LLM, ReAct~\cite{yao2022react}, and
LLM-KBCI~\cite{takayama2024integrating}.
SCD-LLM applies a single LLM pass conditioned on the SCD output (e.g., adjacency
list) and variable meta-data to revise edge existence and/or directions.
ReAct interleaves natural-language reasoning with tool-based actions (e.g., graph
edits and external checks) to iteratively refine the SCD graph.
LLM-KBCI adopts a two-stage ZS-CoT prompting framework: it first elicits
explanations for each (non-)edge and then makes a final discrete decision,
yielding an LLM-refined causal graph.

\section{Instantiation of \texorpdfstring{$\mathrm{Conn}_{\min}$}{Conn\_min} by Path-Cover Closure}
\label{sec:appendix:conn_min}

This appendix describes a practical instantiation of
$\mathrm{Conn}_{\min}(\mathcal{R}\Rightarrow \mathrm{MB}^{\mathcal H}(Y))$ when
we aim to preserve \emph{all} upstream emergence pathways from $\mathcal{R}$ to
each boundary variable. Here, ``$\min$'' denotes the \emph{minimal closure} that
preserves \emph{all} directed paths from $\mathcal{R}$ to every
$B\in \mathrm{MB}^{\mathcal H}(Y)$, i.e., it keeps the union of these paths and
removes any nodes/edges not on any such path.

\paragraph{Key idea.}
We keep exactly the nodes and directed edges that lie on at least one directed
path from any root $r\in\mathcal{R}$ to any boundary variable
$B\in \mathrm{MB}^{\mathcal H}(Y)$ in the (constrained) retained mechanism graph.
Equivalently, the connecting subgraph is the union of \emph{all} directed
$\mathcal{R}\!\to\!B$ paths, which preserves alternative micro/meso mechanisms
leading to the same boundary factor.

\paragraph{Constrained graph.}
Start from the retained mechanism graph $G_W^{(k)}$ and enforce all
simulation-confirmed edges as hard constraints (e.g., fixed orientations and/or
edge inclusion). Let the resulting directed graph be $\widetilde{G}_W^{(k)}$.

\paragraph{Reachability-based path cover.}
Let $\mathrm{Reach}^+(U)$ be the set of nodes reachable from a node set $U$ via
directed edges in $\widetilde{G}_W^{(k)}$, and let $\mathrm{Reach}^-(V)$ be the
set of nodes that can reach a node set $V$ (i.e., reachability in the reverse
graph). For each boundary variable $B$, define
\begin{equation}
\label{eq:path_support}
\mathcal{P}(B)
\;:=\;
\mathrm{Reach}^+(\mathcal{R}) \;\cap\; \mathrm{Reach}^-(\{B\}).
\end{equation}
A node is in $\mathcal{P}(B)$ iff it lies on at least one directed path from
$\mathcal{R}$ to $B$. We then take the union over all boundary variables:
\begin{equation}
\label{eq:path_union}
\mathcal{P}
\;:=\;
\bigcup_{B\in \mathrm{MB}^{\mathcal H}(Y)} \mathcal{P}(B).
\end{equation}
We instantiate $\mathrm{Conn}_{\min}(\mathcal{R}\Rightarrow \mathrm{MB}^{\mathcal H}(Y))$
as the subgraph of $\widetilde{G}_W^{(k)}$ induced by $\mathcal{P}$ (keeping only
directed edges). By construction, this subgraph contains \emph{all} directed
$\mathcal{R}\!\to\!B$ paths that are consistent with the enforced constraints.

\paragraph{Pruning.}
After inducing the subgraph and enforcing hard constraints, we drop any
non-terminal node (excluding $\mathcal{R}$ and $\mathrm{MB}^{\mathcal H}(Y)$)
that is not on a directed path from $\mathcal{R}$ to any boundary variable.
This removes dangling components irrelevant to the emergence explanation.

\paragraph{Result.}
The resulting subgraph serves as
$\mathrm{Conn}_{\min}(\mathcal{R}\Rightarrow \mathrm{MB}^{\mathcal H}(Y))$.
Combined with $\{Y\}\cup \mathrm{MB}^{\mathcal H}(Y)$, it yields the explanatory
subgraph $E_Y$ used to trace micro--meso mechanisms that give rise to $Y$.

\section{O2O Delivery Platform Simulation Details}
\label{app:o2o_simulation}

To comprehensively validate the effectiveness of our method, we construct a multi-agent simulation environment for on-demand food delivery. The simulator models a heterogeneous collaborative ecosystem composed of \emph{Merchants}, \emph{Riders}, and \emph{Customers}. \emph{Importantly, the simulator’s micro-to-macro emergence patterns are explicitly instantiated from ground-truth mechanisms calibrated on real-world data, rather than being assumed implicitly or left to uncontrolled dynamics.}

\subsection{Ecosystem Overview and Interaction Loop}
At each discrete time step $t=1,2,\dots$, the environment evolves through an event-driven pipeline:
\begin{enumerate}
 \item \textbf{Order generation:} Customer agents generate new orders over time following \textbf{empirical trends extracted from real-world data} (e.g., time-of-day demand curves and spatial demand patterns).
 \item \textbf{Merchant processing:} Merchant agents accept and prepare orders, potentially with queueing and preparation delays.
 \item \textbf{Dispatching:} The Dispatch Center assigns available orders to rider agents according to a global policy.
 \item \textbf{Rider response and execution:} Riders respond to dispatch instructions and execute pickup-and-delivery actions (e.g., accept/reject, reroute, reprioritize).
 \item \textbf{Logging:} The simulator records states, actions, outcomes, and derived variables for subsequent causal analysis.
\end{enumerate}
This setup provides a controllable testbed while preserving realistic complexities such as dynamic supply--demand imbalance, congestion, and delayed effects.

\subsection{LLM-Based Cognition-Driven Decision-Making}
Unlike traditional rule-based simulators, core entities in our environment are embedded with a Large Language Model (LLM) core. As a result, an agent's behavior is not a deterministic function mapping from state to action, but a stochastic process driven by semantic understanding and contextual reasoning.

A representative example is the \textbf{Rider agent}. Given (i) its role/persona, (ii) the current \emph{explicit observable state}, and (iii) an \emph{implicit psychological state}, the rider may nonlinearly weigh and respond to dispatch instructions (e.g., accept, reject, delay, or reorder tasks). This design intentionally introduces the real-world \emph{intention--behavior gap}, injecting valuable noise and uncertainty that makes identifying causal chains substantially more challenging and realistic.

\subsection{Multi-Level Dynamic Games and Cascading Effects}
The system reproduces a complex supply--demand game with multi-level decision-making:
\begin{itemize}
  \item The \textbf{Dispatch Center} attempts to allocate orders under a global objective (e.g., latency, throughput, service stability).
  \item Individual \textbf{Riders} respond according to local optimality (e.g., personal revenue, route preferences, workload considerations).
\end{itemize}
The conflict between centralized dispatching and decentralized execution, together with queueing-theoretic stochastic order arrivals, yields a highly coupled dynamical system. Small perturbations at a single node (e.g., a rider rejecting an order or a merchant delaying preparation) can propagate through the system and be nonlinearly amplified via cascading effects, making causal tracing and attribution particularly difficult.

\subsection{Micro-Interactions, Macro-Emergence, and Dynamic Latent State Space}
Macroscopic system performance (e.g., average delivery time, overtime rate, backlog, stability) is not determined by any single variable; instead, it emerges from nonlinear interactions among a large number of micro-level entities.

To operationalize and quantitatively detect such emergence, following the value-entropy-based indicator proposed by \citet{yu2025unlocking}, we define the emergence score $Y$ as the degree to which the system deviates from its optimal operating regime due to excessive disorder, measured by an edge value-entropy loss:
\begin{equation}
Y = 1 - \exp\!\left(-\frac{H_T - H_{\text{best}}}{H_{\text{best}}}\right).
\end{equation}
Here, $H_T$ denotes the system entropy at time step $t$, computed from the empirical probability distribution of \emph{order contract deviations}, i.e., the difference between the actual delivery time and the promised delivery time, which reflects whether an order is delivered early, on time, or late. $H_{\text{best}}$ is the optimal reference entropy corresponding to the best operating efficiency of the platform (in our calibration, $H_{\text{best}}\approx 1.1609$). When $Y=0$, the system stays within an ordered (or mildly fluctuating) regime where the vast majority of orders can be delivered as expected, and the overall macro-state is controllable and predictable. As $Y$ increases and approaches $1$, the system exhibits severe disorder characterized by large-scale delays or capacity collapse, indicating an emergence regime where normal scheduling becomes ineffective and service quality turns highly unstable and difficult to predict.

\subsection{Experimental Configuration and Scalability}
In our experiments, we use a default configuration with \textbf{12 merchants}, \textbf{41 customers}, and \textbf{20 riders}. All population sizes are configurable, allowing us to scale the number of merchants/users/riders and to modify load conditions (e.g., demand intensity and service capacity) for robustness evaluation.

\clearpage
\section{Prompt Templates for CAMO Agents}
\label{app:prompt_templates}





\subsection{A1: Worldview Parser}

\begin{PromptBoxA}{A1: System Prompt}
You construct a structured, computable worldview from "fragmented facts".
You must follow the Global Contract. When {fragmented_facts} are insufficient to support recurring patterns
or latent mechanisms, you must retrieve supporting literature using available tools before answering.
\end{PromptBoxA}

\begin{PromptBoxA}{A1: Task 1 --- Multi-Stakeholders and Resources/Constraints/Goals}
Extract multi-stakeholders relevant to {req} from {fragmented_facts}.
For each stakeholder, list resources, constraints, and goals. Keep items concise and actionable.

JSON:
{
  "stakeholders": [
    {
      "name": "string",
      "resources": ["string"],
      "constraints": ["string"],
      "goals": ["string"]
    }
  ]
}
\end{PromptBoxA}

\begin{PromptBoxA}{A1: Task 2.1 --- Behavior Patterns (with automatic evidence retrieval)}
Identify recurring behavior patterns that appear across multiple independent sources or multiple cases.
Express each pattern as an If...Then... rule (or equivalent).
If multi-source/multi-case support is not clear from {fragmented_facts}, retrieve empirical/survey evidence via LIT_SEARCH.
Attach evidence ids to each pattern (from retrieved items or directly from {fragmented_facts} if traceable).

JSON:
{
  "patterns": [
    {
      "if_clause": "string",
      "then_clause": "string",
      "drivers": ["string"],
      "tag": "multi_sources | multi_cases",
      "evidence_ids": ["string"]
    }
  ]
}
\end{PromptBoxA}

\begin{PromptBoxA}{A1: Task 2.2 --- Latent Factors (ground mechanisms with tools when needed)}
Derive latent motivations/mechanisms behind the extracted behavior patterns.
If mechanisms are not directly supported by {fragmented_facts}, retrieve literature evidence (LIT_SEARCH) and,
when necessary, verify details via PAPER_FETCH.
Label each factor as "directly observed", "inferred", or "severely underspecified".
Attach evidence ids for each factor.

JSON:
{
  "latent_factors": [
    {
      "name": "string",
      "description": "string",
      "label": "directly observed | inferred | severely undersspecified",
      "evidence_ids": ["string"]
    }
  ]
}
\end{PromptBoxA}

\begin{PromptBoxA}{A1: Task 3 --- Multi-Scale Structural Relationships (Micro/Meso/Macro)}
Construct multi-scale structures (micro/meso/macro) from {fragmented_facts} and {latent_variables}.
For each scale, provide one structural assumption and map latent variables (by name) to exactly one scale.

JSON:
{
  "micro": {"structural_assumption": "string", "mapping": ["string"]},
  "meso":  {"structural_assumption": "string", "mapping": ["string"]},
  "macro": {"structural_assumption": "string", "mapping": ["string"]}
}
\end{PromptBoxA}

\begin{PromptBoxA}{A1: Task 4 --- Consistency and Stability Measurement}
Compute the intersection and union sizes between two sets.
Treat semantically equivalent elements as identical when counting.

JSON:
{
  "intersection_size": "int",
  "union_size": "int"
}
\end{PromptBoxA}


\subsection{A2: Worldview Integrator}

\begin{PromptBoxB}{A2: System Prompt}
You unify and resolve conflicting worldviews arising from multiple perspectives, sources, and explanations.
You must follow the Global Contract.
\end{PromptBoxB}

\begin{PromptBoxB}{A2: Task 1 --- Language Unification of Indicators}
Deduplicate latent variables with the same semantics but different names.
Align statistical calibers and provide a unique definition for each unified indicator.

JSON:
{
  "unified_indicators": [
    {
      "name": "string",
      "description": "string",
      "calculation_formula": "string",
      "data_source": "string",
      "time_window": "string"
    }
  ]
}
\end{PromptBoxB}

\begin{PromptBoxB}{A2: Task 2 --- Determine Variable Links}
Determine association relationships among variables in {unified_latent_variables} using {behavior_patterns}.
All variables must be selected from {unified_latent_variables}.
If multiple relationships exist for the same pair, include all with distinct descriptions.

JSON:
{
  "variable_links": [
    {
      "name": "string",
      "targets": [{"name": "string", "description": "string"}]
    }
  ]
}
\end{PromptBoxB}

\begin{PromptBoxB}{A2: Task 3 --- Maintain Competing Explanations}
For each (source,target) link, provide competing explanations that differ materially.
Include prerequisites, evidence, and a support estimate in [0,1].
Output must include every link, even if only one explanation exists.

JSON:
{
  "competition_relationship": [
    {
      "source": "string",
      "target": "string",
      "explanations": [
        {"text": "string", "prerequisites": "string", "evidence": "string", "support_estimation": "float"}
      ]
    }
  ]
}
\end{PromptBoxB}

\begin{PromptBoxB}{A2: Task 4 --- Rating of Candidate Causal Graph}
Rate the causal graph from three aspects in [0,100]: fit, simplicity, explainability.

JSON:
{"fit":"int","simplicity":"int","explainability":"int"}
\end{PromptBoxB}

\begin{PromptBoxB}{A2: Task 5 --- Reassess Uncertain Variables (Data Feedback)}
Rejudge whether uncertain variables should remain, using {gt_feedback} and A3's reasons.
Set is_important=false for at most 5 least reliable variables; set true for all others.

JSON:
{
  "uncertain_variable_reassess": [
    {"name":"string","is_important":"bool","reason":"string"}
  ]
}
\end{PromptBoxB}

\begin{PromptBoxB}{A2: Task 6 --- Latent Variable Regeneration}
Redesign variables flagged for improvement (definition/formula/source/window).
Optionally add up to 2 new variables (must not duplicate {certainty_latent_variables}).

JSON:
{
  "regenerated_uncertain_indicators": [
    {"name":"string","description":"string","calculation_formula":"string","data_source":"string","time_window":"string"}
  ]
}
\end{PromptBoxB}


\subsection{A3: Causal Cartographer}

\begin{PromptBoxC}{A3: System Prompt}
You assemble candidate causal structures by aligning worldview variables with dataset constraints,
then annotate edges with identifiability and produce feedback.
You must follow the Global Contract. When causal discovery methods, identifiability criteria,
or orientation rules require external evidence, you must retrieve literature (LIT_SEARCH / WEB_SEARCH)
and verify key details (PAPER_FETCH) before finalizing outputs.
\end{PromptBoxC}

\begin{PromptBoxC}{A3: Task 1 --- Node Adaptation}
For each latent variable, decide whether it is observable, constructible, or uncertain given {data_columns}.
If constructible, provide a formal formula using only names from {data_columns}.
Variables in {certain_variables_names} must be retained as observable or constructible.

JSON:
{
  "variable_examination": [
    {"name":"string","flag":"observable | constructible | uncertain","explanation":"string","formula":"string"}
  ]
}
\end{PromptBoxC}

\begin{PromptBoxC}{A3: Task 1.5 --- Feedback to A2}
Provide concise guidance (<=200 words) on accepted/rejected variables and how to improve variable generation.

JSON:
{
  "feedback_to_A2":"string",
  "accepted_variables":["string"],
  "rejected_variables":[{"name":"string","reason":"string"}]
}
\end{PromptBoxC}

\begin{PromptBoxC}{A3: Task 2 --- Causal Structure Generation (Merge + Tool-Triggered Method Lookup)}
Merge worldview-driven and data-driven structures into a candidate structure.
Use node adaptation formulas: if a variable is constructible, add implied edges consistent with its formula.
If you need to choose or justify a causal discovery strategy (e.g., PC/GES/LiNGAM/NOTEARS/FGES/FCI, etc.),
or need rules about identifiability under assumptions, you MUST retrieve evidence via LIT_SEARCH/WEB_SEARCH first,
and optionally PAPER_FETCH for verification. Record evidence ids in each edge if such justification is used.

JSON:
{
  "candidate_structure": [
    {"source":"string","target":"string","description":"string","support_estimation":"float","flag":"world | data | both",
     "evidence_ids":["string"]}
  ]
}
\end{PromptBoxC}

\begin{PromptBoxC}{A3: Task 3 --- Edge Identifiability Analysis (tool-using when criteria are external)}
Add an identifiability label to each edge: identifiable, assumption-dependent, or non-identifiable.
If your identifiability decision relies on formal criteria beyond the given structure/assumptions, retrieve references
via LIT_SEARCH/WEB_SEARCH and verify with PAPER_FETCH when needed. Attach evidence ids when used.

JSON:
{
  "edge_identifiability": [
    {"source":"string","target":"string","description":"string","support_estimation":"float","flag":"world | data | both",
     "identifiability_flag":"identifiable | assumption-dependent | non-identifiable",
     "evidence_ids":["string"]}
  ]
}
\end{PromptBoxC}

\begin{PromptBoxC}{A3: Task 3.5 --- Counterfactual Edge Orientation}
Determine the edge direction based on counterfactual experimental results.
If you invoke any external orientation principle (e.g., invariance assumptions, SCM criteria),
retrieve evidence first and attach evidence ids.

JSON:
{
  "oriented":"bool",
  "direction":"source_to_target | target_to_source | bidirectional | undetermined",
  "confidence":"float",
  "reasoning":"string",
  "evidence_source_to_target":"string",
  "evidence_target_to_source":"string",
  "evidence_ids":["string"]
}
\end{PromptBoxC}

\begin{PromptBoxC}{A3: Task 4 --- Feedback Iteration (Blind Spots)}
Identify which parts of outcome_variable remain poorly explained by the current graph.
Return feedback items as hypotheses for missing variables/structures.

JSON:
{"feedback_items":["string"]}
\end{PromptBoxC}

\subsection{A4: Simulation Scriptwright.}

\begin{PromptBoxD}{A4: System Prompt}
You design structured simulation experiment packages to test important and uncertain causal edges.
You must follow the Global Contract.
\end{PromptBoxD}

\begin{PromptBoxD}{A4: Task 1 --- Edge Importance Evaluation}
Add an importance score in [0,1] for each edge, reflecting sensitivity to the objective in {req},
and provide a brief explanation.

JSON:
{
  "edge_importance_res": [
    {"source":"string","target":"string","description":"string","support_estimation":"float","flag":"string",
     "identifiability_flag":"string","importance":"float","importance_explanation":"string"}
  ]
}
\end{PromptBoxD}

\subsection{A5: Counterfactual Adjudicator}

\begin{PromptBoxE}{A5: System Prompt}
You verify causal graphs using counterfactual simulation results: evaluate credibility, design interventions,
verify edges, and refine graphs. You must follow the Global Contract.
\end{PromptBoxE}

\begin{PromptBoxE}{A5: Task 1 --- Model Credibility Evaluation}
Compare simulation output with real-world observations and return a credibility score in [0,1].
If credibility is low, treat downstream findings as in-model evidence only.

JSON:
{"credibility_score":"float","explanation":"string"}
\end{PromptBoxE}

\begin{PromptBoxE}{A5: Task 2 --- Experiment Design}
Design default settings and multiple intervention settings by varying target_variable while keeping others fixed.

JSON:
{
  "default_values":{"var_name":"value"},
  "experiments":[{"experiment_id":"string","description":"string","settings":{"var_name":"value"}}]
}
\end{PromptBoxE}

\begin{PromptBoxE}{A5: Task 3 --- Causal Edge Verification}
For each edge, assign one label:
"In-model strong support" | "In-model partial support" | "In-model refutation" | "Insufficient evidence".

JSON:
[
  {"source":"string","target":"string","verification_label":"string","reasoning":"string"}
]
\end{PromptBoxE}

\begin{PromptBoxE}{A5: Task 4 --- Refine Causal Graph}
Remove refuted edges, keep supported edges, and decide how to treat insufficient-evidence edges.

JSON:
[
  {"source":"string","target":"string","relation":"string"}
]
\end{PromptBoxE}

\section{Licenses and Terms}
\label{app:license}

This work uses third-party artifacts (models, software libraries, and public aggregate statistics). We summarize the corresponding licenses/terms, usage, and redistribution status in Table~\ref{tab:licenses}. We do not redistribute any third-party proprietary artifacts (e.g., API-served model weights); all third-party resources are accessed and used under their original licenses/terms.

Our released code and any derived artifacts (e.g., prompts and configurations) are intended solely for research and reproducibility in controlled simulation settings. We do not claim or recommend direct real-world deployment without domain-specific validation and appropriate compliance and ethical review.

\paragraph{Privacy note.}
Real-world data are used only in non-identifying, aggregate form for simulator calibration (Table~\ref{tab:licenses}); we do not collect, store, or release any personally identifiable information. All analyses are performed on synthetic simulation rollouts.

\begin{table*}[t]
\centering
\small
\caption{Licenses and terms for artifacts used in this work.}
\setlength{\tabcolsep}{5pt}
\renewcommand{\arraystretch}{1.15}
\begin{tabular}{p{2.9cm} p{3.3cm} p{3.2cm} p{2.8cm} c}
\toprule
\textbf{Artifact} & \textbf{Provider / Source} & \textbf{License / Terms} & \textbf{How used} & \textbf{Re-dist.} \\
\midrule

\multicolumn{5}{l}{\textit{LLM backbones}}\\
Qwen3-235B-A22B &
Qwen (official release) &
Apache License 2.0 &
LLM backbone &
No \\

DeepSeek-R1-Distill-Qwen-32B &
DeepSeek (official release) &
MIT License &
LLM backbone &
No \\

DeepSeek-V3.2 &
DeepSeek (official release) &
MIT License &
LLM backbone &
No \\

GPT-5 mini &
OpenAI API &
OpenAI Terms of Use + Service Terms
+ Usage Policies
+ Data controls
&
LLM backbone (API) &
No \\

Gemma3-27B &
Google DeepMind &
Gemma Terms of Use &
LLM backbone &
No \\
\midrule
\addlinespace[2pt]
\multicolumn{5}{l}{\textit{Data / statistics}}\\
Meituan Research~\citep{wang2025meituan} aggregate statistics &
Meituan-INFORMS-TSL Research Challenge repo &
CC BY-NC 4.0&
Simulator calibration (aggregate only) &
No \\
\midrule
\addlinespace[2pt]
\multicolumn{5}{l}{\textit{Software dependencies}}\\
SCD library (PC/FCI/GES/MMHC implementation) &
causal-learn/pgmpy &
MIT License &
SCD baselines &
No \\
\bottomrule
\end{tabular}
\label{tab:licenses}
\end{table*}

\section{Computational Budget and Infrastructure}
\label{app:budget}

We report the computing infrastructure and approximate resource usage for the experiments. Hardware details are summarized below, and the average LLM API consumption per trial is reported in Table~\ref{tab:api_consumption}.

\subsection{Computing Infrastructure}
All local computations (multi-agent coordination, environment state updates, and data processing) were executed on a dedicated server:
\begin{itemize}
  \item \textbf{CPU:} Dual AMD EPYC 9654 (96 cores, 2.4\,GHz per CPU), totaling 192 physical cores.
  \item \textbf{GPU:} $4 \times$ NVIDIA RTX A6000 (48\,GB per card; 192\,GB total).
  \item \textbf{RAM:} 512\,GB DDR5 ECC ($16 \times 32$\,GB; 4800\,MHz).
  \item \textbf{Software:} Ubuntu 22.04 LTS and Python-based simulation frameworks.
\end{itemize}

\subsection{GPU Hours and Parallelization}
GPUs were used for lightweight model inference, while CPUs handled discrete-event simulation logic and agent state management. The total GPU usage was approximately 2.5 GPU-hours.

\subsection{LLM API Consumption}
Table~\ref{tab:api_consumption} reports the average number of API calls and tokens per trial, where $N$ denotes the number of major iterative loops in the simulation.

\begin{table*}[t]
  \centering
  \small
  \setlength{\tabcolsep}{10pt}
  \renewcommand{\arraystretch}{1.15}
  \caption{Average LLM API consumption per experimental trial.}
  \label{tab:api_consumption}
  \begin{tabular}{lcc}
    \toprule
    \textbf{Task Type} & \textbf{API Calls / Run} & \textbf{Tokens / Run} \\
    \midrule
    Method (Core Logic)      & $50$--$80$  & $350{,}000$--$1{,}000{,}000$ \\
    Simulation (Environment) & $70$--$150$ & $(200{,}000$--$400{,}000)\times N$ \\
    \bottomrule
  \end{tabular}

  \vspace{2pt}
  {\footnotesize \textit{Note:} $N$ denotes the number of major iterative loops in the simulation.}
\end{table*}

\section{Parameters for Packages}
\label{app:pkg-params}

We report third-party packages used for causal discovery, preprocessing, and (partial) LLM inference, together with the key parameter settings in Table~\ref{tab:pkg-params}.

\begin{table*}[t]
  \centering
  \small
  \setlength{\tabcolsep}{6pt}
  \renewcommand{\arraystretch}{1.15}
  \caption{Packages and key parameter settings.}
  \label{tab:pkg-params}
  \begin{tabular}{p{0.20\textwidth} p{0.35\textwidth} p{0.30\textwidth}}
    \hline
    \textbf{Package / Algorithm} & \textbf{Implementation} & \textbf{Key parameters} \\
    \hline
    pgmpy / PC &
    pgmpy.estimators.PC &
    significance\_level=0.05 \\
    
    pgmpy / GES &
    pgmpy.estimators.HillClimbSearch &
    scoring\_method=BicScore(df), max\_iter=1e4 \\
    
    pgmpy / MMHC &
    pgmpy.estimators.MmhcEstimator &
    default (estimate()) \\
    
    causal-learn / FCI &
    causallearn.search.ConstraintBased.FCI.fci &
    chisq test, alpha=0.05 \\
    
    scikit-learn / Discretization &
    sklearn.preprocessing.KBinsDiscretizer &
    n\_bins=5, encode=ordinal, strategy=quantile \\
    
    vLLM / LLM inference (partial) &
    vLLM serving/decoding &
    temperature=0.7, quantization=AWQ, context=131072, max tokens=32768 \\
    \hline
  \end{tabular}
\end{table*}
\end{document}